\documentclass[lettersize,journal]{IEEEtran}
\usepackage{amsmath,amsfonts}
\usepackage{algorithmic}
\usepackage{algorithm}
\usepackage{array}
\usepackage{textcomp}
\usepackage{stfloats}
\usepackage{url}
\usepackage{verbatim}
\usepackage{graphicx}
\usepackage{cite}
\usepackage{amssymb}
\usepackage{threeparttable}
\usepackage{multirow}
\usepackage[table]{xcolor}
\usepackage{booktabs}
\usepackage{subfigure}
\usepackage{amsmath,bm}
\usepackage{orcidlink}
\begin{document}

\title{3D Lane Detection from Front or Surround-View using Joint-Modeling \& Matching}

\author{Haibin Zhou~\orcidlink{0009-0003-8322-5888}, Huabing Zhou~\orcidlink{0000-0001-5007-7303}, \textit{Member, IEEE}, Jun Chang~\orcidlink{0000-0002-3594-9879}, Tao Lu~\orcidlink{0000-0001-8117-2012}, \textit{Member, IEEE}, and Jiayi Ma~\orcidlink{0000-0003-3264-3265}, \textit{Senior Member, IEEE}
\thanks{This work was supported in part by the National Natural Science Foundation of China under Grant 62171327,62171328 and 62072350,  and in part by Key R\&D Program in Hubei Province, China under Grant 2022BAA079. \textit{(Corresponding author: Huabing Zhou.)}}
\thanks{Haibin Zhou, Huabing Zhou, and Tao Lu are with the Hubei Key Laboratory of Intelligent Robot, Wuhan Institute of Technology, Wuhan 430205, China.
        {\tt\small (e-mail: csidez@163.com, zhouhuabing@gmail.com, lutxyl@gmail.com)}}%
\thanks{Jun Chang is with the Computer Science School, Wuhan University, Wuhan 430072, China.
        {\tt\small (e-mail: chang.jun@whu.edu.cn)}}%
\thanks{Jiayi Ma is with the Electronic Information School, Wuhan University, Wuhan 430072, China.
        {\tt\small (e-mail: jyma2010@gmail.com)}}
}

\markboth{IEEE Transactions on Intelligent Vehicles}%
{Shell \MakeLowercase{\textit{et al.}}: A Sample Article Using IEEEtran.cls for IEEE Journals}


\maketitle

\begin{abstract}
3D lanes offer a more comprehensive understanding of the road surface geometry than 2D lanes, thereby providing crucial references for driving decisions and trajectory planning. While many efforts aim to improve prediction accuracy, we recognize that an efficient network can bring results closer to lane modeling. However, if the modeling data is imprecise, the results might not accurately capture the real-world scenario. Therefore, accurate lane modeling is essential to align prediction results closely with the environment. This study centers on efficient and accurate lane modeling, proposing a joint modeling approach that combines B\'{e}zier curves and interpolation methods. Furthermore, based on this lane modeling approach, we developed a Global2Local Lane Matching method with B\'{e}zier Control-Point and Key-Point, which serve as a comprehensive solution that leverages hierarchical features with two mathematical models to ensure a precise match. We also introduce a novel 3D Spatial Encoder, representing an exploration of 3D surround-view lane detection research. The framework is suitable for front-view or surround-view 3D lane detection. By directly outputting the key points of lanes in 3D space, it overcomes the limitations of anchor-based methods, enabling accurate prediction of closed-loop or U-shaped lanes and effective adaptation to complex road conditions. This innovative method establishes a new benchmark in front-view 3D lane detection on the Openlane dataset and achieves competitive performance in surround-view 2D lane detection on the Argoverse2 dataset.
\end{abstract}


\section{Introduction}
\IEEEPARstart{L}{ane} detection research \cite{tang2021review, bar2014recent} in autonomous driving \cite{chen2022milestones} has made significant advancements \cite{ranft2016role}. To address the automotive industry's needs, it has transitioned from predicting lanes in image space \cite{lu2021super} to predicting lanes in bird's-eye view (BEV) perspectives \cite{ma2022vision}, and 3D space. Compared to BEV (which represents lanes in 2D space) \cite{gosala2022bird, reiher2020sim2real}, 3D lanes enable a deeper understanding of the road's geometric structure. For example, constructing high-precision maps using 3D lane detection can help prevent misalignment and overlap issues between elevated and ground-level roads.

\begin{figure}[t]
\centering{
\subfigure[Front-View Image]{
\label{Fig.sub.1}
\includegraphics[width=1.6in]{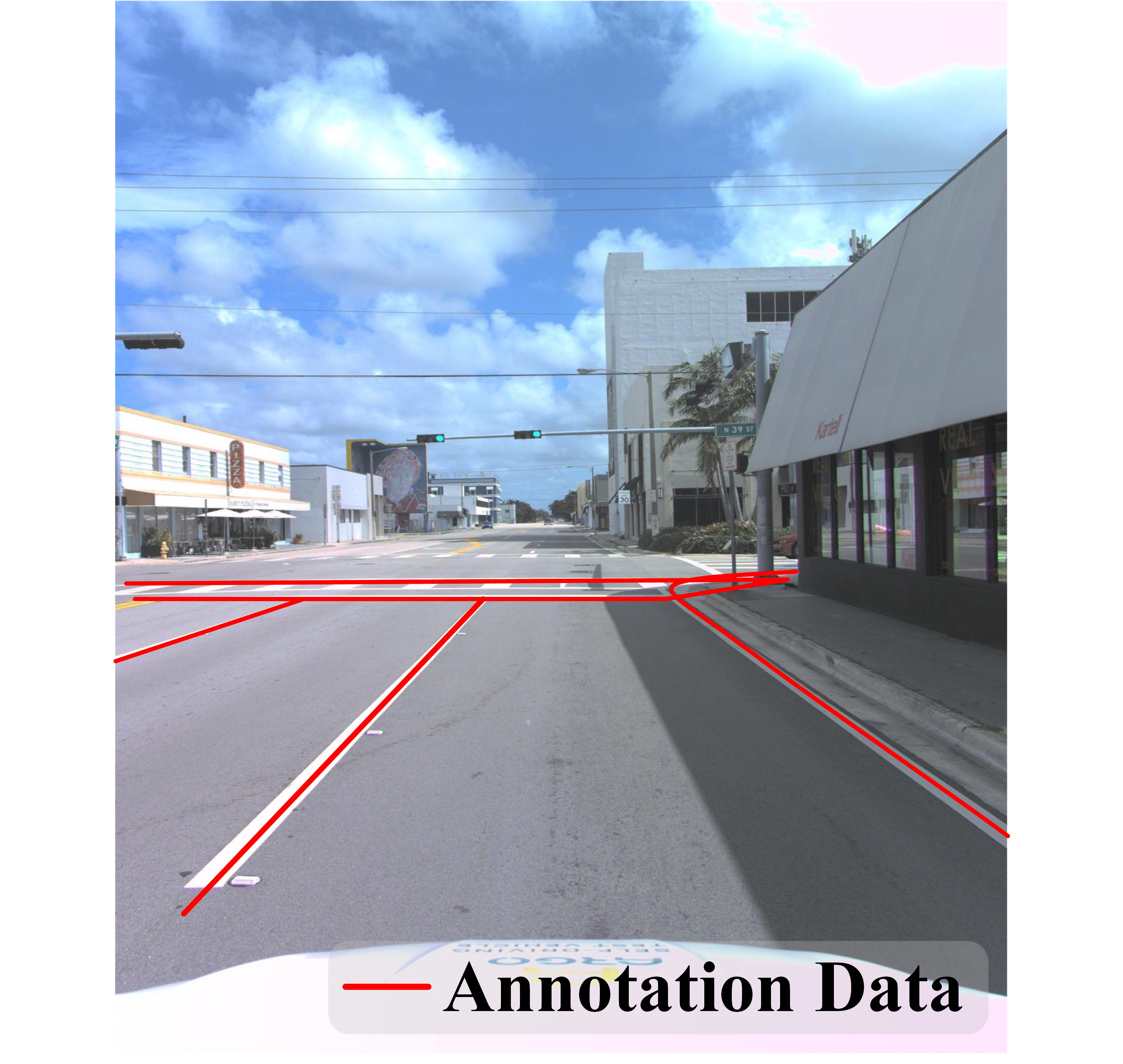}}
\subfigure[Polynomial Curve]{
\label{Fig.sub.2}
\includegraphics[width=1.6in]{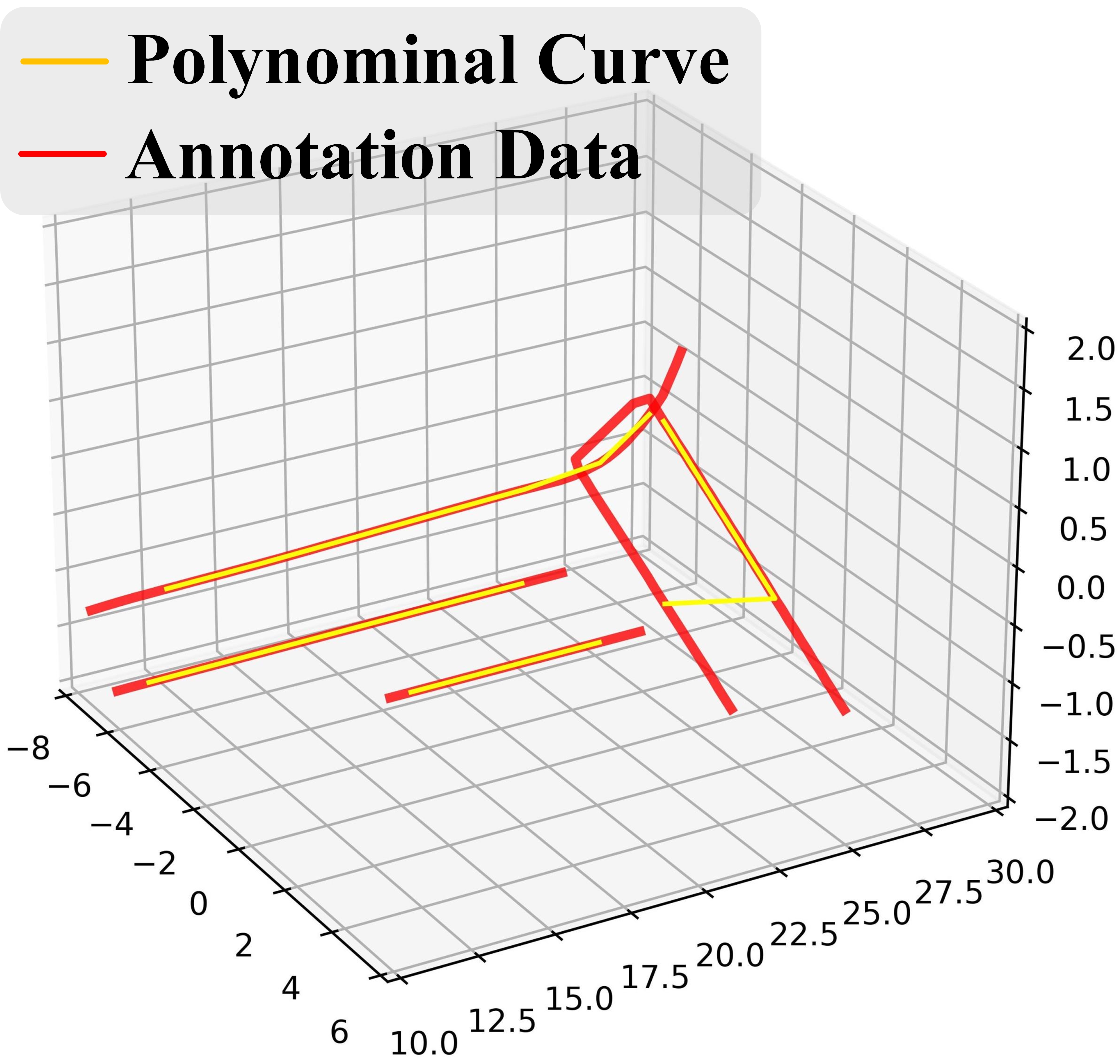}}
\subfigure[Interpolation Curve]{
\label{Fig.sub.3}
\includegraphics[width=1.6in]{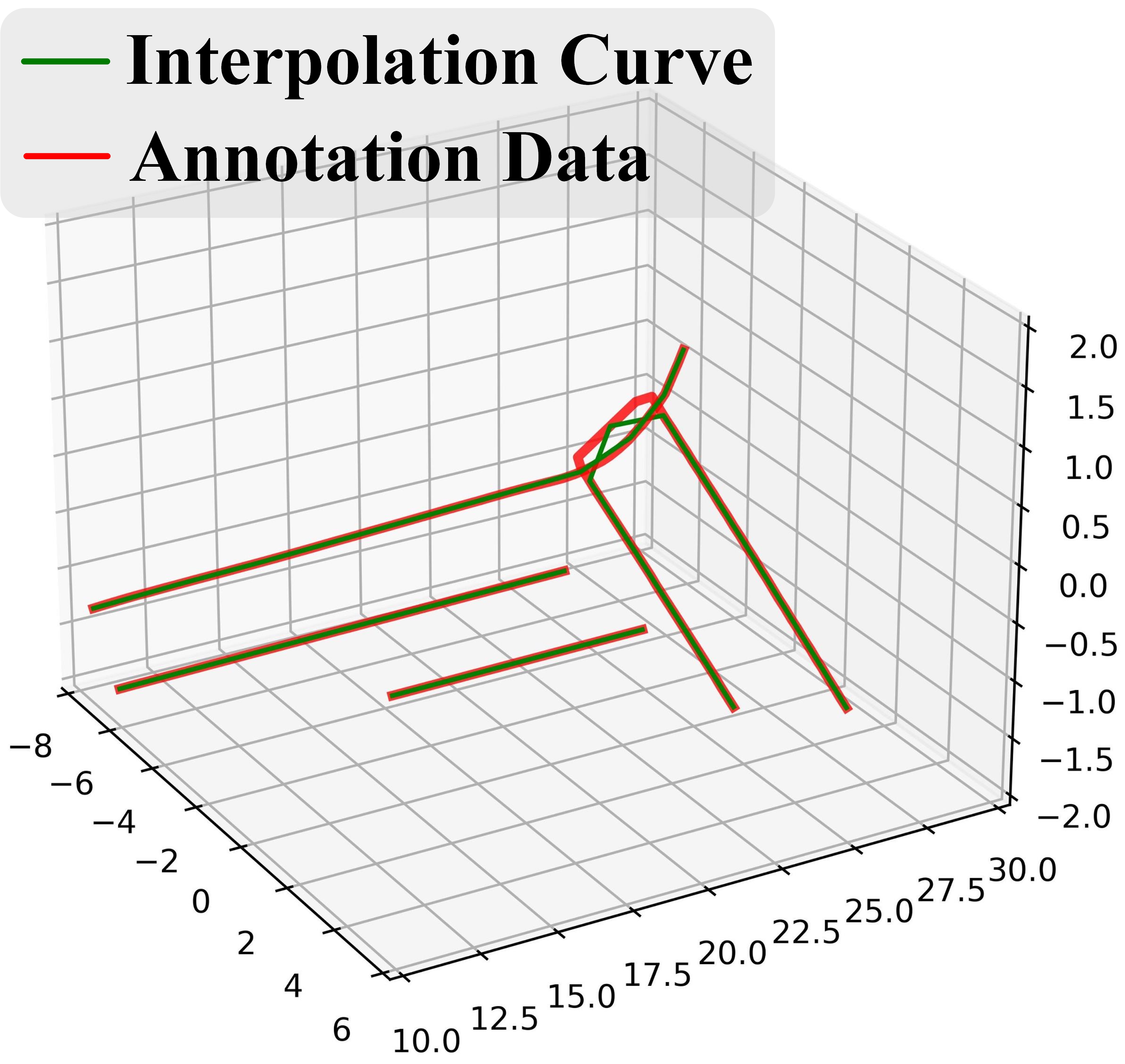}}
\subfigure[B\'{e}zier Curve with Control Points]{
\label{Fig.sub.4}
\includegraphics[width=1.6in]{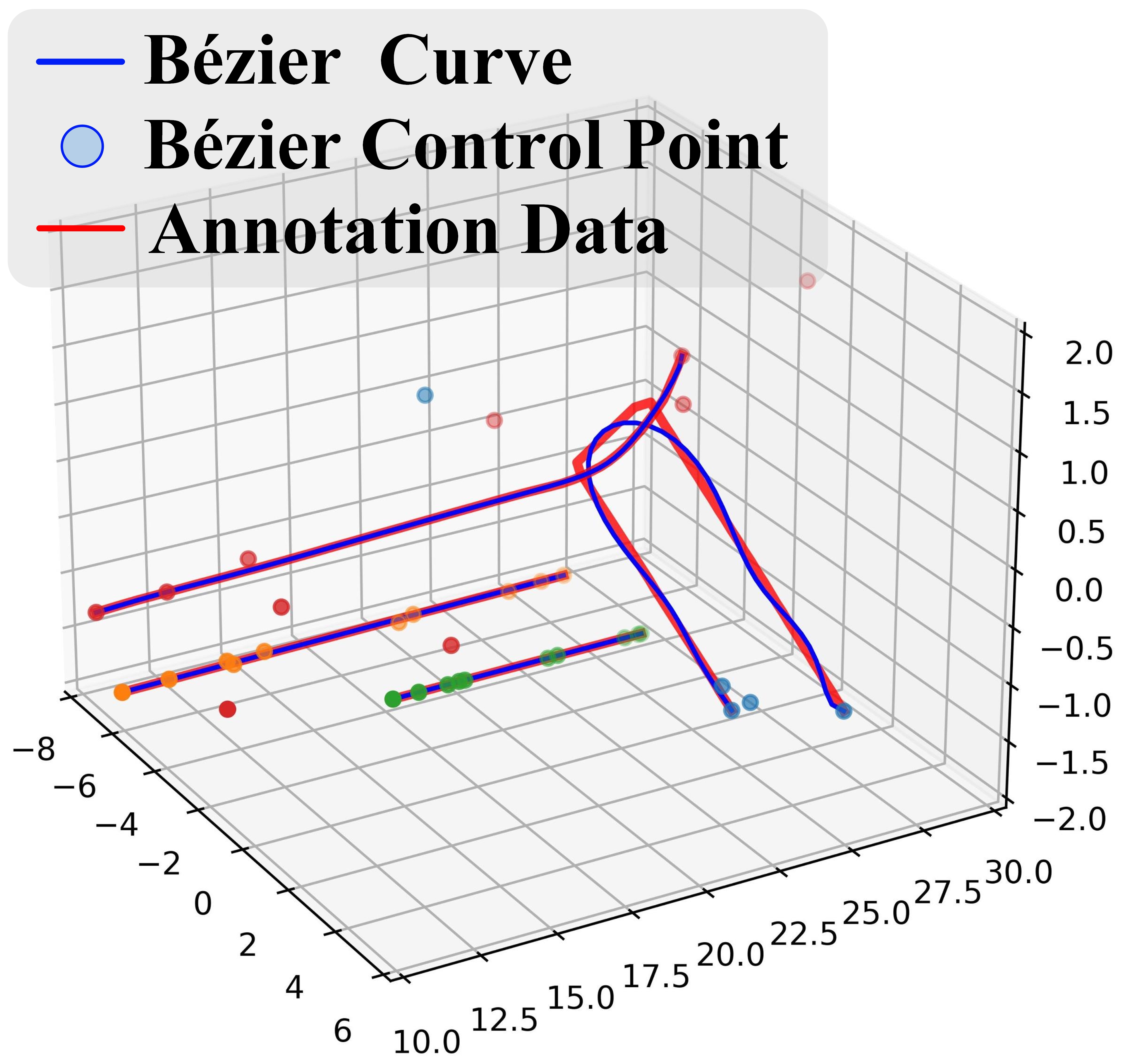}}
\caption{Illustration of the impact of three modeling methods on lanes in front-view.
\textbf{Polynomial Curve}: This method involves two polynomial functions, one describing the X-Y relationship and another for the Z-Y relationship.
\textbf{Interpolation Curve}: This approach is based on interpolation and represented by a fixed set of key points.
\textbf{B\'{e}zier Curve with Control Points}: This method yields smooth and accurate curves by manipulating control points. (A set of points sharing the same color governs the same lane line.)}
\label{fig_1}}
\vspace{-0.4cm}
\end{figure}

In 2D or 3D lane detection work \cite{sun2022adaptive, li2022hdmapnet, liao2022maptr, liu2023vectormapnet, bai2023curveformer, chen2022persformer, garnett20193d}, a common practice involves transforming lane annotation data of various numbers and lengths into a standardized mathematical model. Some methods \cite{chen2022persformer, bai2023curveformer, garnett20193d} achieve 3D lane modeling by constructing the X-Y and Z-Y function relationships using polynomials \cite{feng2022rethinking}. The accuracy of these modeling methods is dependent on the dimensions of the polynomial (referred to as the traditional \textbf{Polynomial Curve}). In contrast, several methods \cite{li2022hdmapnet, liao2022maptr, liu2023vectormapnet} describe a 2D lane using a limited set of key points derived from interpolation (referred to as the \textbf{Interpolation Curve}), which potentially yields more precise lane modeling compared to the polynomial curve. Motivated by this, we employed this approach to compute a fixed set of key points to represent the lane lines in 3D space. However, modeling a highly curved line necessitates a greater number of key points \cite{liu2020abcnet}. \IEEEpubidadjcol To address this issue, we discovered that B\'{e}zier control points can depict a lane line using fewer points, while also encompassing information about the line's curvature (referred to as the \textbf{B\'{e}zier Curve with Control Points}). Lanes modeled using the B\'{e}zier method exhibit smoother characteristics. We illustrate the modeling effects of the three methods from a front-view perspective in Fig.~\ref{fig_1}. Fig.~\ref{Fig.sub.3} and Fig.~\ref{Fig.sub.4} confirm that both the Interpolation Curve and the B\'{e}zier Curve with Control Points can accurately characterize lanes in the images.

Currently, most front-view 3D lane detection methods \cite{garnett20193d, guo2020gen, chen2022persformer, bai2023curveformer} use the Polynomial Curve modeling method due to their anchor-based design, which computes offsets from initial points on the Y-axis to fit ground truth. While this approach effectively handles straight lines in forward-view cameras, real-world lanes often display more complex patterns, such as loops (indicating the U-shaped curve) and closed loops (enclosing flowerbeds). On the other hand, recent 2D surround-view lane detection methods \cite{liao2022maptr, liu2023vectormapnet} use the Interpolation Curve for modeling. This method predicts key points on the lane to construct vectorized high-definition maps \cite{li2022hdmapnet}. Despite its ability to predict complex curves through key point detection, the inherent limitations of the Interpolation Curve modeling result in insufficient smoothness of curved segments. Consequently, while the predicted results may approximate the modeled lane, they do not precisely correspond to the annotated data. In this paper, we employ both the Interpolation Curve and the B\'{e}zier Curve methods for lane line modeling, an approach that considers both the smoothness and accuracy of the lines. We design a \textbf{G}lobal2\textbf{L}ocal Lane Matching mechanism with \textbf{B}\'{e}zier Control-Point and \textbf{K}ey-Point (\textbf{GL-BK}). The GL-BK lane matching mechanism adeptly directs the network to predict the 3D coordinates of lane key points in terms of category, position, shape, and smoothness using geometric information from the B\'{e}zier Control-Point and Key-Point, which integrates the supervision of two mathematical models.

We have observed that many networks use Polynomial Curve modeling output curve parameters. These networks need to set a fixed polynomial order, which leads to a limitation in representation, making them unsuitable for handling scenarios with complex curves. Inspired by 2D vectorized maps that predict key points, this approach's high flexibility suits various scenarios. Our network processes 2D images from front-view or surround-view cameras, outputting lane line key points in 3D space. What's more, in order to transform the 2D features extracted by the backbone into 3D space, we further developed the 3D Spatial Encoder, which adopts a voxel-like query \cite{mao2021voxel} to generate 3D features by sampling each 3D reference point.

Our method has achieved state-of-the-art performance on the front-view public dataset Openlane \cite{chen2022persformer}. It has also successfully implemented surround-view 3D lane detection on Argoverse2 \cite{wilson2023argoverse}, demonstrating excellent performance. Furthermore, leveraging Argoverse2, we generated a set of Front-View datasets Argoverse2$^{\dag}$ that contain more complex lanes compared to Openlane, showcasing the performance of this framework.

In summary, the contributions of our method are as follows:
\begin{itemize}
	\item We focus on lane modeling and propose a novel 3D lane modeling approach which combines B\'{e}zier curves and interpolation methods.
	\item We design a hierarchical lane matching mechanism, GL-BK, which leverages multiple features and mathematical models to ensure a precise match between predicted lane lines and their ground truths.
	\item We develop a novel 3D Spatial Encoder for point-level 3D lane detection, addressing anchor-based method limitations.
	\item The framework achieves state-of-the-art performance in predicting 3D front-view lanes and competitive results in 2D surround-view lane prediction, marking it as the first 3D surround-view lane detection network.
\end{itemize}

\section{Related Work}
\subsection{Lane Line Modeling}
In the lane detection task, the initial step involves the modeling of lanes \cite{liu2023vectormapnet}, which entails converting each lane line into a fixed-scale vector. Presently, there are two principal approaches for lane line modeling. One approach is based on linear interpolation \cite{blu2004linear}, a method employed to create smooth, continuous lines between given coordinate points. By computing the slopes between adjacent points, additional intermediate points can be inserted to minimize variability. Subsequently, a fixed number of key points are selected along the smooth line. MapTR \cite{liao2022maptr} is a method based on linear interpolation, utilizing key points to model 2D lane annotations. Alternatively, another approach is based on curve fitting, which adjusts the smoothness of the curve by appropriately selecting the order of the curve to fit the ground truth and achieve modeling. Feng et al. \cite{feng2022rethinking} proposed a method that uses the B\'{e}zier curve-based method to model lanes in image space. Persformer \cite{chen2022persformer} employs the polynomial curve-based method to model 3D lanes, which is also the most commonly used method for 3D lane modeling. Accurate modeling of annotated data ensures that the constructed ground truth accurately mirrors the layout of lanes in real-world scenarios. In contrast, incorrect modeling may not accurately reflect the real environment, even if the network's predicted results closely approach the modeled ground truth.

\subsection{2D Lane Detection}
The 2D lane detection task is divided into two categories. The first category includes traditional lane detection methods that detect lanes in front-view images. LaneATT \cite{tabelini2021keep} uses an anchor-based \cite{zhang2020bridging} model to balance accuracy and efficiency in lane detection in front-view images. LSTR \cite{liu2021end} uses a transformer-based network to predict the parameters of a lane shape model. GANet \cite{wang2022keypoint} treats lane detection as key-point estimation. Feng et al. \cite{feng2022rethinking} propose a parametric B\'{e}zier curve-based method for lane detection in image space. The second category involves inputting front-view or surround-view images and outputting lane detection results in BEV. HDMapNet \cite{li2022hdmapnet} constructs a local semantics lane map in BEV based on multiple sensor observations. VectorMapNet \cite{liu2023vectormapnet} is an end-to-end vectorized High-definition (HD) map that predicts sparse polylines in BEV using key-point detection on lanes. MapTR \cite{liao2022maptr} introduces a unified permutation-equivalent modeling approach and is the first method to achieve real-time and high-precision vectorized HD map construction.

\subsection{3D Front-View Lane Detection}
In recent years, the attention towards 3D lane detection has significantly grown compared to 2D. This is mainly due to the limitations of 2D lanes, which lack depth information and susceptibility to error propagation in spatial transformations. Several methods have made notable progress in this field. 3D-LaneNet \cite{garnett20193d} employs a dual-pathway architecture that integrates inverse perspective mapping within the network and utilizes an anchor-based lane representation. By projecting features into a bird's-eye view space using inverse perspective mapping and adopting an anchor-based approach, it effectively represents lanes. Building upon this, 3D-LaneNet+ \cite{efrat20203d} divides BEV features into distinct non-overlapping grids and detects lanes by regressing lateral offset distance, line angle, and height offset relative to each grid's center. This method is capable of detecting complex lane topology. Gen-LaneNet \cite{guo2020gen} introduces a novel geometry-guided lane anchor representation method that employs a virtual bird's-eye view coordinate system instead of the vehicle coordinate system. It directly calculates the 3D lane points output by the network through specific geometric transformations. PersFormer \cite{chen2022persformer} constructs dense BEV queries with known camera poses and optimizes anchor design to provide better feature representation, achieving unified 2D and 3D lane detection. CurveFormer \cite{bai2023curveformer} employs a one-stage transformer-based method to directly predict 3D lane parameters. It uses curve queries and iteratively refines anchor points to approach the ground truth. To the best of our knowledge, no exploration has been done for 3D surround-view lane detection methods.

\section{Method}
\subsection{Joint Lane Modeling as Ground-Truth for Training}
Lane modeling is a crucial data preprocessing step within the entire training process. We represent a single lane line using key points based on linear interpolation and B\'{e}zier control points derived from B\'{e}zier curves.

Linear interpolation, a linear function, has the advantage of low computational complexity. Its key points provide a straightforward way to determine the local position of lanes. However, it may lead to accuracy loss when dealing with complex curves or a large number of data points. Hence, B\'{e}zier curves may be more suitable in such situations.

On the other hand, the B\'{e}zier curve-based method offers excellent smoothness and can adjust the curve's shape by controlling the position and number of control points, resulting in a continuous and smooth curve. However, as demonstrated by the U-shaped curve in Fig.~\ref{Fig.sub.4}, higher-order B\'{e}zier curves may exhibit oscillations when modeling certain simple curve shapes.

In conclusion, while linear interpolation is particularly effective for modeling simple curves, B\'{e}zier curves stand out when it comes to representing complex curves. Therefore, we propose the concept of Joint Lane Modeling to combine these two methods for more comprehensive lane modeling. Mathematically, this can be represented as $\xi_\text{frame} = {{{\bm C} \cup {\bm K}}}$. Here, $\xi_\text{frame}$ represents the lanes within a frame. The term ${\bm C} \in \mathbb{R}^{(L\times P_{c} \times 3)}$ signifies $L$ curves controlled by $P_c$ three-dimensional control points, which is used as ground truth. Similarly, ${\bm K} \in \mathbb{R}^{(L\times P_{k} \times 3)}$ denotes the same L curves,  but represented using $P_k$ three-dimensional coordinate key points. The following provides a detailed explanation of how to generate ${\bm K}$ and ${\bm C}$.

\textbf{Preliminaries on B\'{e}zier Control Points.} The B\'{e}zier curve is a parametric mathematical curve. B\'{e}zier control points are the key points that define the B\'{e}zier curve. These points are not always directly on the curve, but they determine the shape and direction of the curve. The definition and shape of the B\'{e}zier curve are dependent on these control points. 

B\'{e}zier curves and traditional polynomial curves are both parametric mathematical curves. However, B\'{e}zier curves hold a distinct advantage. Adjusting a single control point allows for the strongest influence on the curve's shape near the corresponding region of that control point's parameter value, diminishing as one moves away. This stands in contrast to polynomial curves, where a change in any coefficient could impact the entire curve. Therefore, B\'{e}zier curves are better suited for fitting complex and varied lane lines. 

Based on the standard B\'{e}zier formula, the B\'{e}zier curve $\mathcal{B}(t)$ can be represented by control points $\mathcal{C}$ and Bernstein basis polynomials $b_{n}(t)$ as follows:
\vspace*{-0.3 \baselineskip}
\begin{equation}\label{eq:03}
\mathcal{B}(t) = \sum_{j=0}^{n}  b_{j, n}(t) \mathcal{C}_j, \ t \in [0, 1],
\vspace*{-0.3 \baselineskip}
\end{equation}
where $n$ denotes the order of the B\'{e}zier curve, and $n+1$ represents the number of control points.
In this paper, we calculate a lane line as an example ($L = 1$). We aim to describe a lane line as a B\'{e}zier curve, represented as a set of annotated data points ${\mathcal{A} = \{(x_i, y_i, z_i)\}_{i=0}^{P_a-1}}$, and obtain control points $\mathcal{C} = \{(x_j, y_j, z_j)\}_{j=0}^{P_c-1}$, where $P_a$ is the number of annotated data points and $P_c$ is the number of control points. With the following formula, $t_i$ achieves the uniform distribution of annotated data points in the interval $[0, 1]$:
\vspace*{-0.3 \baselineskip}
\begin{equation} \label{eq:02}
t_i=\frac{i}{P_a-1}, \ i \in \{0, 1, 2, ..., P_a-1\}.
\vspace*{-0.3 \baselineskip}
\end{equation}
Therefore, we use $t_i$ to express the function $\mathcal{A}(t_i)$ as follows:
\vspace*{-0.3 \baselineskip}
 \begin{equation}\label{eq:01}
\mathcal{A}(t_i) = \sum_{j=0}^{n}  b_{j, n}(t_i) \mathcal{C}_j, \ n=P_c-1.
\vspace*{-0.3 \baselineskip}
\end{equation}

For Bernstein basis polynomials, $b_{j, n}(t_i)$ provides different weights for the control points at different values of the parameter $t_i$, thus offering control over the shape of the curve. We express $b_{j, n}(t_i)$ using the following equation:
\vspace*{-0.3 \baselineskip}
\begin{equation}\label{eq:00}
b_{j, n}(t_i) = \frac{n!}{j! \times n!} \times (1-t_i)^{n-j} \times (t_i)^{j}.
\vspace*{-0.3 \baselineskip}
\end{equation}

In the following, we adopt matrix notation to simplify Eq.~\eqref{eq:01}:
\vspace*{-0.3 \baselineskip}
\begin{equation}\label{eq:1}
\begin{bmatrix} \!A{(t_0)}\!\\  \!A{(t_1)}\!\\ \!\vdots\! \\ \! A{(t_{p_a-1})}\!\end{bmatrix} \!=\!
\begin{bmatrix}  \!b_{0,n}(t_0)\!&\!\cdots\!&\! b_{p_c-1,n}(t_0)\!\\ \!b_{0,n}(t_1)\!&\!\cdots\!&\! b_{p_c-1,n}(t_1)\!\\ \!\vdots\!&\!\ddots\!&\!\vdots\!\\  \!b_{0,n}(t_{p_a-1})\!&\!\cdots\!&\! b_{p_c-1,n}(t_{p_a-1})\!\end{bmatrix}\!
\begin{bmatrix} \! C_0\!\\ \! C_1\!\\ \!\vdots\!\\ \!C_{p_c-1}\!\end{bmatrix}\!,
\vspace*{-0.3 \baselineskip}
\end{equation}
\vspace*{-0.3 \baselineskip}
\begin{equation}\label{eq:2}
\bm A = \bm{b} \bm{C},
\vspace*{-0.3 \baselineskip}
\end{equation}
where $\bm A$ is the vector consisting of $P_a$ sample points on the B\'{e}zier curve, $\bm{b}$ is a matrix with each row representing the Bernstein basis functions at the parameter $t_i$, and $\bm C$ is the vector of $P_c$ control points.

Then, we use the least squares method to compute $\bm{C}$, aiming to minimize the residual $||\bm A-\bm b \bm{C}||_F^2$. Thus, the solution to minimize this expression is:
\vspace*{-0.3 \baselineskip}
 \begin{equation}\label{eq:3}
\bm C = (\bm{b}^T \bm{b})^{-1}\bm{b}^T \bm{A}.
\vspace*{-0.3 \baselineskip}
\end{equation}

\begin{figure*}[t]
\centering
\includegraphics[width=7in]{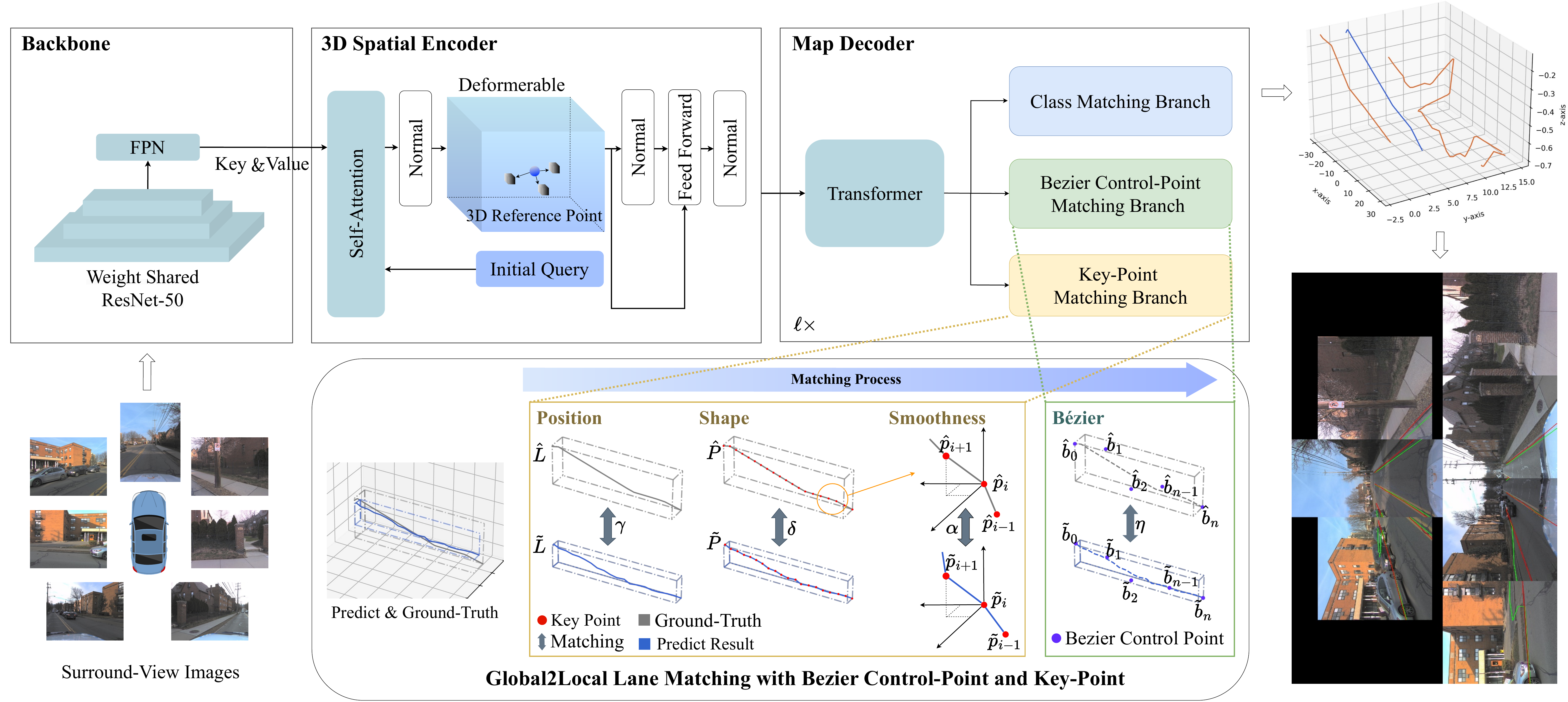}
\caption{Model overview. As illustrated, the network processes surround-view images and produces the 3D coordinates of key points for each lane, along with their respective categories. The 3D Spatial Encoder utilizes a voxel-like query to elevate the 2D image features extracted by the Backbone to 3D space. The Transformer of Map Decoder output key points feature ${\bm Q}_k$ and control points feature ${\bm Q}_c$, and feeds them to three matching branches. These three matching branches conduct Global2Local Lane Matching with B\'{e}zier Control-Point and Key-Point. For detailed information about ${\bm Q}_k$ and ${\bm Q}_c$, please refer to the Map Decoder under the 3D Lane Detection Network section. (The network architecture for front-view image is consistent with that of surround-view.)}
\label{fig_2}
\vspace{-0.4cm}
\end{figure*}

\textbf{Preliminaries on Fixed-Num Key Point.} Linear interpolation is used to estimate the values of points between two given points. Since the number of keys on each line in labeled data $\bm A$ is different, we may need to employ interpolation to pad the key points in order to convert $\bm A$ into a fix-scale ${\bm K}$. Consider two points, ${A}_1 = ({x}_{1}, { y}_{1}, { z}_{1})$ and ${ A}_2 = ({ x}_{2}, { y}_{2}, { z}_{2})$. Using linear interpolation, we can calculate the interpolated point $\mathcal{K}(t)$ between them, where $t$ is a parameter that ranges from 0 to 1. The equation for 3D linear interpolation is expressed as:
\vspace*{-0.3 \baselineskip}
\begin{equation}  \label{eq:4}
 \mathcal{K}(t) = (1 - t) \times { A}_1 + t \times { A}_2.
 \vspace*{-0.3 \baselineskip}
\end{equation}

When $t = 0$, $\mathcal{K}(t)$ equals ${ A}_1$ and when $t = 1$, $  \mathcal{K}(t)$ equals ${ A}_2$. As $t$ varies within the range $[0, 1]$, $  \mathcal{K}(t)$ gradually transitions from ${A}_1$ to ${A}_2$, achieving a smooth interpolation between the two points.

\vspace{-0.1cm}
\subsection{3D Lane Detection Network}
The network architecture, illustrated in Fig.~\ref{fig_2}, supports input from surround-view or front-view images and predicts 3D lanes around or in front of the ego vehicle. Importantly, the processing method is the same for both input types. Therefore, the following discussion will be presented using only the surround-view input as an example. This architecture produces real-world lane coordinates using the ego vehicle as the origin, precisely mapped into image space via the perspective projection matrix.  The network comprises three main components: Backbone, 3D Spatial Encoder, and Map Decoder. 

\textbf{Backbone.} The Backbone accepts image inputs from various perspectives, each sharing a ResNet-50 \cite{he2016deep} with identical parameters. This step produces multi-scale features for each perspective at different stages. These features are then fed into the Feature Pyramid Network (FPN) \cite{lin2017feature} to aggregate comprehensive environmental information and are subsequently reshaped as Key and Value for the following components. The purpose of the FPN is to improve the network's ability to detect features of distant, small objects. 

\textbf{3D Spatial Encoder.} To convert 2D features from image space to 3D space, we design the 3D Spatial Encoder. This module generates voxel-like 3D features that represent the 3D space. The BEVFormer \cite{li2022bevformer}, based on Deformable Attention \cite{zhu2020deformable}, proposes the Spatial Cross-Attention using BEV queries to obtain BEV features. Inspired by the design of the Spatial Cross-Attention module, we designed our 3D Spatial Encoder. However, in contrast to the BEVFormer, which uses a pillar-like query to sample multiple 3D reference points from a pillar to represent a BEV feature, the 3D Spatial Encoder adopts a voxel-like query. Each query samples a single 3D reference point, outputting a voxel-like feature that preserves height information. While the BEVFormer produces a final output without height-related 2D sampled features, the 3D Spatial Encoder provides 3D sampled features for generating 3D lanes. Here, the self-attention focuses on the relationships across different perspectives.

\begin{figure}[t]
\centering
\includegraphics[width=3.5in]{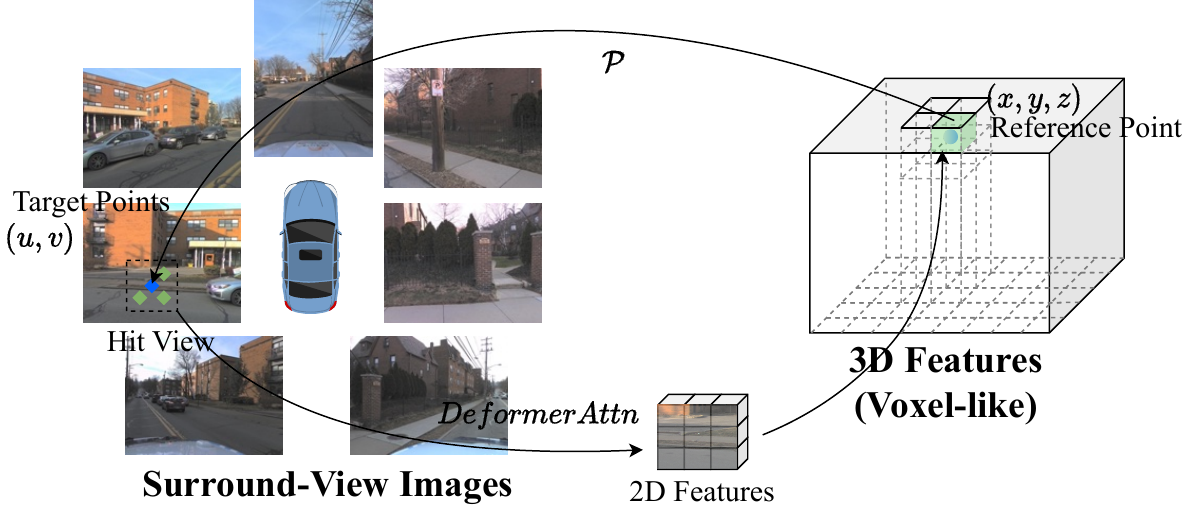}
\caption{The process of the query to generate 3D features. For a voxel-like query, the 3D reference point $(x, y, z)$ is projected to the 2D point $(u, v)$ with the function $\mathcal{P}$. Based on the Deformable Attention (DeformAttn), sampling is performed around the target point $(u, v)$ to obtain a weighted sum of 2D features. We project the 2D features back into a corresponding voxel in the 3D grid. After traversing the entire 3D grids, comprehensive 3D features are generated.}
\label{fig_3}
\vspace{-0.4cm}
\end{figure}

As illustrated in Fig.~\ref{fig_3}, we first construct an empty 3D grid based on the visible range and then define the number of divisions for the x, y, and z axes to segment the 3D grid. Next, we use the camera's intrinsic and extrinsic parameters to build the projection matrix $\bm M$ for the 3D-to-2D view transformation. For each voxel-like query, a 3D reference point $(x, y, z)$ is sampled and projected to its corresponding point $(u, v)$ in the 2D view via the matrix ${\bm M}$. This process is formulated as follows:
\vspace*{-0.3 \baselineskip}
\begin{equation} \label{eq:5}
(u, v)_{n} =  \mathcal{P}((x, y, z), {\bm M}_n), \ n \in \mathcal{V}_{hit}.
\vspace*{-0.3 \baselineskip}
\end{equation}

Note that $n$ is the index of the hit view, in this projection process the number of hit views $V_{hit}$ may exceed one. To extract relevant features from the hit views as the results of voxel-like queries ${\bm Q}_{3d}$, we choose Deformable Attention proposed by Deformable DETR\cite{zhu2020deformable} to reduce computational cost. The Deformable Attention samples only a few points around the reference point in the input feature map as attention keys. Based on Deformable Attention, we generate a set of the most relevant points around the point $(u, v)$. These learned points in the 2D view are defined as target points. Subsequently, we compute the weighted sum of these sampled 2D features, producing features corresponding to the voxel-like query represented as ${\bm F}_{voxel}$, formulated as:
\vspace*{-0.3 \baselineskip}
\begin{equation} \label{eq:6}
{\bm F}_{voxel} = \frac{1}{|\mathcal{V}_{hit}|} \sum_{n \in \mathcal{V}_{hit}} {\text{DeformAttn}} ({\bm Q}_{3d}, (u, v)_{n} , {\bm F}^{n}),
\vspace*{-0.3 \baselineskip}
\end{equation}
where $\mathcal{P}$ is the projection function, which uses the projection matrix ${\bm M}$ to project the reference points to corresponding points in hit view. ${\bm F}^{n}$ represents the $n$-th image features where the target point is located.

As the query iterates over the entire 3D grid, a comprehensive 3D feature representation is generated, denoted as 3D features $\bm {F}_{3d}$.

\textbf{Map Decoder.} The decoder comprises two sets of queries and $\ell$ numbers of transformer decoder layers, each passing through three matching branches. The decoder outputs the last of the $\ell$ layers as the best-performing prediction result. 

We define a set of Key-Point queries, where the query results are regarded as expanded key point features $\bm Q_{k} \in \mathbb{R}^{(\hat{L}\times P_{k} \times N)}$. Here, $\hat{L}$ represents the max number of predicted lane lines in the current scene, $P_{k}$ represents the number of key points in each lane, and $N$ represents the channel number. In the \textit{Key-Point Matching Branch}, we extract 3D spatial key points $\hat{\bm K} \in \mathbb{R}^{(\hat{L} \times P_{k} \times 3)}$ from ${\bm Q}_{k}$, where the $3$ representing the three dimensions of the predicted lane's key point coordinates. The \textit{Class Matching Branch} extracts the category information $\hat{\bm c} \in \mathbb{R}^{(\hat{L} \times P_{o})}$ of each lane line from ${\bm Q}_{k}$, where $P_{o}$ represents the total number of categories. 

Additionally, we define a set of B\'{e}zier Control-Point queries, which yield control point features ${\bm Q}_{c} \in \mathbb{R}^{(\hat{L} \times P_c \times N)}$. The \textit{B\'{e}zier Control-Point Matching Branch} outputs the control points $\hat{\bm C} \in \mathbb{R}^{(\hat{L} \times P_c \times 3)}$, where $P_{c}$ represents the number of control points we use to describe a B\'{e}zier curve.

The \textit{Key-Point Matching Branch} and \textit{B\'{e}zier Control-Point Matching Branch} achieve coordinate regression, while the \textit{Class Matching Branch} performs classification. Each branch consists of three layers: Linear, Layer Normal, and Rule layers. Each matching branch implements its own matching rules, as described in the GL-BK Lane Matching section.
\vspace{-0.1cm}

\subsection{GL-BK Lane Matching}
In our proposed network, given a single-frame input, it predicts lanes as $\hat{\bm{K}} \in \mathbb{R}^{(\hat{L} \times P_k \times 3)}$ and $\hat{\bm C} \in \mathbb{R}^{(\hat{L} \times P_c \times 3)}$. The ground truth is denoted as ${\bm K} \in \mathbb{R}^{({L} \times P_k \times 3)}$ and ${\bm C} \in \mathbb{R}^{({L} \times P_c \times 3)}$, where ${L} \leqslant \hat{L}$. Here, we use $\varnothing$ (no object) to pad the ground truth from ${L}$ to $\hat{L}$, forming $\tilde{\bm K} \in \mathbb{R}^{(\hat{L} \times P_k \times 3)}$ and $\tilde{\bm C} \in \mathbb{R}^{(\hat{L} \times P_c \times 3)}$. And ${\bm c}$ is the ground truth of the classification information. The objective is to find a permutation of $\hat{L}$ elements in $\hat{\bm K}$ with the minimal matching cost to prepare for loss calculations. We present a GL-BK Lane Matching method. The process is formulated as:
\vspace*{-0.3 \baselineskip}
\begin{equation} \label{eq:7}
\begin{aligned}
\hat{\pi} &= \underset {\pi \in \prod_{L}} {\text{argmin}} \sum_{i=0}^{L-1} ( \lambda \mathcal{L}_{position}(\hat{\bm K}_{\pi(i)}, \tilde{\bm K}_{i})
\\ &\quad+ \alpha \mathcal{L}_{shape}(\hat{\bm K}_{\pi(i)}, \tilde{\bm K}_{i}) + \beta \mathcal{L}_{smoothness}(\hat{\bm K}_{\pi(i)}, \tilde{\bm K}_{i})
\\ &\quad+ \gamma \mathcal{L}_{Bezier}(\hat{\bm C}_{\pi(i)}, \tilde{\bm C}_{i}) + \delta \mathcal{L}_{class}(\hat{\bm c}_{\pi(i)}, {\bm c}_{i}) ),
\end{aligned}
\vspace*{-0.3 \baselineskip}
\end{equation}
where $\pi$ denotes a permutation of $L$ lane lines within a single frame, while $\hat{\pi}$ represents the permutation of $L$ lane lines with the lowest matching cost. The function $\pi(i)$ returns the $i$-th element in the permutation $\pi$. The parameters $\lambda, \alpha, \beta, \gamma$, and $\delta$ indicate the weights of the five respective matching costs. The GL-BK Lane Matching method uses B\'{e}zier control points to capture both the positional information and local curvature of lanes for a comprehensive global matching of curves. Moreover, key points of the curves ensure precise alignment concerning position, shape, and smoothness.

\textbf{Key-Point Matching in Position.} We describe the contour and position of a lane line in 3D space using a 3D bounding box. The Key-Point queries produce the predicted lane, $\hat{\bm K}$. Selecting a lane line from $\hat{\bm K}$ is denoted as $\hat{\bm K}_{\pi(i)} \in \mathbb{R}^{(P_k \times 3)}$, with $\hat{\bm K}_{\pi(i)} = \{(x_{n}, y_{n}, z_{n})\}$, where $n \in [0 , P_k-1]$. By computing the maximum and minimum values across the X, Y, and Z dimensions, we define the lane line bounding box as $\hat{\bm B} \in \mathbb{R}^{(L \times 6)}$ with $\hat{\bm B}_{\pi(i)} = (x_{min}, y_{min}, z_{min}, x_{max}, y_{max}, z_{max})$. Similarly, select a lane line $\tilde{K}_i$ from $\tilde{K}$, and calculate its bounding box as $\tilde{\bm B_i}$. A pairwise match between $\hat{\bm B}_{\pi(i)}$ and $\tilde{\bm B}_{i}$ is conducted using L1 Loss \cite{girshick2015fast}. The following equation, where $\mathcal{F}_{bt}$ denotes a bounding box transformer function converting a point sequence in lane lines to bounding boxes, represents the process:
\vspace*{-0.3 \baselineskip}
\begin{gather} \label{eq:8_9}
\hat{\bm B}_{\pi(i)} = \mathcal{F}_{bt}(\hat{\bm K}_{\pi(i)}), \\
\mathcal{L}_{position}(\hat{\bm K}_{\pi(i)}, \tilde{\bm K}_{i}) = \mathcal{L}_{L1}(\hat{\bm B}_{\pi(i)}, \tilde{\bm B}_{i}).
\vspace*{-0.3 \baselineskip}
\end{gather}

\textbf{Key-Point Matching in Shape.} By matching the key points within the lanes, We refine the shape of the curves and partially constrain the local curvature. we employ the Euclidean distance pair-wise between $\hat{\bm K}$ and $\tilde{\bm K}$ to calculate the matching cost. The process is formulated as follows:
\vspace*{-0.3 \baselineskip}
\begin{gather} \label{eq:10}
\mathcal{L}_{shape}(\hat{\bm K}_{\pi(i)}, \tilde{\bm K}_{i})\! =\! \frac{1}{P_k} \sum_{j=0}^{P_k-1} \mathcal{L}_{Euclidean}(\hat{\bm K}_{\pi(i)}^{j}, \tilde{\bm K}_{i}^{j}),
\vspace*{-0.3 \baselineskip}
\end{gather}
where the $\hat{\bm K}_{\pi(i)}^{j}$ represents the $j$-th point on the $\pi(i)$-th lane lines. Note that the $P_k$ represents the key points number in each lane.

\textbf{Key-Point Matching in Smoothness.} Curvature is a geometric property that describes the amount of bending or deviation of a curve or surface from a straight line \cite{roberts2001curvature}. In 3D space, the curvature of a lane line can be calculated using the cosine value on the lane.
We calculate the tangent vector $\hat{\bm T}_j$ on each point $\hat{\bm K}_{\pi(i)}^{j}=(x_j, y_j, z_j)$ as follows:
\vspace*{-0.3 \baselineskip}
\begin{equation} \label{eq:11}
\hat{\bm T}_j = (x_{j+1} - x_{j-1}, y_{j+1} - y_{j-1}, z_{j+1} - z_{j-1}),
\vspace*{-0.3 \baselineskip}
\end{equation}
where $0 < j < P_k-1$.

We estimate the local curvature of the $j$-th point on the $\pi(i)$-th lane, which lines in the predicted result $\hat{\bm K}$, defined as $\hat{\bm \Gamma}_{\pi(i)}^j$:
\vspace*{-0.3 \baselineskip}
\begin{equation} \label{eq:12}
\hat{\bm \Gamma}_{\pi(i)}^j = \frac{|\hat{\bm T}_j - \hat{\bm T}_{j-1}|}{|(x_j, y_j, z_j) - (x_{j-1}, y_{j-1}, z_{j-1})|}.
\vspace*{-0.3 \baselineskip}
\end{equation}
Then we calculate matching cost of smoothness as follows:
\vspace*{-0.3 \baselineskip}
\begin{gather} \label{eq:13}
\mathcal{L}_{smoothness}(\hat{\bm K}_{\pi(i)}, \tilde{\bm K}_{i}) = \frac{1}{P_k-1} \sum_{j=1}^{P_k-2} \mathcal{L}_{L1}(\hat{\bm \Gamma}_{\pi(i)}^j, \tilde{\bm \Gamma}_{\pi(i)}^j).
\vspace*{-0.3 \baselineskip}
\end{gather}

\textbf{Control-Point Matching in B\'{e}zier.} The B\'{e}zier curves described by B\'{e}zier key points can fit the annotated data well, enabling a comprehensive global matching of the curves. The matching of control points is also a crucial step in matching costs. Given the predicted control points $\hat{\bm C}_{\pi(i)}$, and their corresponding ground truth $\tilde{\bm C}_{i}$, the B\'{e}zier curve is obtained by computing the weighted average:
\vspace*{-0.3 \baselineskip}
\begin{gather} \label{eq:14}
\mathcal{L}_{Bezier}(\hat{\bm C}_{\pi(i)}, \tilde{\bm C}_{i}) = \frac{1}{P_c} \sum_{j=0}^{P_c-1} \mathcal{L}_{Euclidean}(\hat{\bm C}_{\pi(i)}^j, \tilde{\bm C}_{i}^j).
\vspace*{-0.3 \baselineskip}
\end{gather}

\textbf{Class Matching.} In class matching cost term, we use the Focal Loss \cite{lin2017focal} to calculate between predicted classification score $\hat{\bm{c}}_{\pi(i)}$ and target class label ${\bm c}_{i}$. We formula the process as follows:
\vspace*{-0.3 \baselineskip}
\begin{gather} \label{eq:15}
\mathcal{L}_{class}(\hat{\bm c}_{\pi(i)}, {\bm c}_{i}) = \mathcal{L}_{Focal}(\hat{\bm c}_{\pi(i)}, {\bm c}_{i}).
\vspace*{-0.3 \baselineskip}
\end{gather}

\vspace{-0.1cm}
\subsection{Loss}
After determining the optimal instance-level assignment $\hat{\pi}(i)$ using the Hungarian algorithm as in DETR, we compute the loss. The following formula employed for this loss computation retains the same components and weights as used in the matching cost calculation:
\vspace*{-0.3 \baselineskip}
\begin{equation}  \label{eq:16}
\begin{aligned}
\mathcal{L}_{3D} &=  \lambda \mathcal{L}_{position}(\hat{\bm K}_{\hat{\pi}(i)}, \tilde{\bm K}_{i}) + \alpha \mathcal{L}_{shape}(\hat{\bm K}_{\hat{\pi}(i)}, \tilde{\bm K}_{i})
\\ &+ \beta \mathcal{L}_{smoothness}(\hat{\bm K}_{\hat{\pi}(i)}, \tilde{\bm K}_{i}) + \gamma \mathcal{L}_{Bezier}(\hat{\bm C}_{\hat{\pi}(i)}, \tilde{\bm C}_{i})
\\ &+ \delta \mathcal{L}_{class}(\hat{\bm c}_{\hat{\pi}(i)}, {\bm c}_{i}).
\end{aligned}
\vspace*{-0.3 \baselineskip}
\end{equation}

\section{Implementation Details}
\vspace{-0.1cm}
\subsection{Datasets}
Openlane and Argoverse2 are public datasets used in autonomous driving, which employ LiDAR and cameras for data collection. Lidar features include 3D point cloud sequences that support 3D object annotation and reconstruction. Therefore, both the Argoverse2 and Openlane datasets directly provide annotated data about 3D lane line key points from LiDAR point clouds. In this paper, we use the 3D lane key point annotations provided by public datasets for lane modeling and network training.

\textbf{Openlane.} Openlane is currently the most comprehensive real-world \textbf{\textit{3D Front-View}} lane dataset, comprising 200K frames with over 880K rigorously annotated lanes. Derived from the Waymo Open dataset, it provides lane and nearest-in-path object annotations across 1000 segments, providing a valuable resource for the development and evaluation of 3D perception algorithms as well as autonomous driving systems. Our study uses 798 segments provided for training and 202 segments for validation.

\textbf{Argoverse2.} We use Argoverse2 to train our \textbf{\textit{3D Surround-View}} lane detection model. Developed by Argo AI, Argoverse2 is an open-source evolution of the original Argoverse. It is an updated version of the original Argoverse dataset, featuring a larger scale and more diverse scenes. This dataset contains high-resolution sensor data such as LiDAR and camera images, along with precise vehicle localization and trajectory information. Covering various urban driving scenarios, including highways, city streets, and complex intersections, Argoverse2 aims to assist researchers and developers in algorithm research and performance evaluation for autonomous driving. Additionally, it provides benchmark tasks and evaluation tools to foster advancements in autonomous driving technology.

\textbf{Argoverse2$^{\dag}$.} This derivative dataset was crafted to offer \textbf{\textit{3D Front-View}} annotations rooted in Argoverse2. While the Openlane Dataset is replete with sample data for 3D front-view tasks, it falls short in providing annotations for horizontal lines perpendicular to the front view. Predominantly, lines in a front-view perspective extend forward, however, there are also many horizontal lines perpendicular to the front-view perspective or other complex curves (e.g., closed loops and loops) in the real world. Argoverse2 provides surround-view 3D lane annotation, including diverse lane line data types such as transverse lines on sidewalks and closed-loop lines on flowerbed boundaries. To overcome Openlane's restriction to single curve types, we leveraged the sensor's intrinsic and extrinsic parameters, combined with the vehicle's pose data, to generate a tailored set of 3D lane data for 3D front-view tasks. 

To generate 3D front-view lanes from the 3D surround-view lanes ground truth, we conduct a systematic procedure outlined in Algorithm~\ref{algorithm:a1}. The main process involves projecting the surround-view 3D ground truth into the front-view camera's 2D image space. Next, we apply a visibility range filter in the image space. Finally, we project back into the 3D space to obtain the front-view 3D lane ground truth.

\begin{algorithm}
\caption{A transformation from 3D surround-view lanes to 3D front-view lanes} \label{algorithm:a1}
\begin{algorithmic} [1]
\renewcommand{\algorithmicrequire}{\textbf{Input:}}
\renewcommand{\algorithmicensure}{\textbf{Output:}}
\REQUIRE{A set of points on 3D surround-view lanes ground truth $\mathcal{A}_{sv}$, front-view intrinsic $\bm{K}$, front-view extrinsic $\bm{T}$, image size dimensions $h, w$.}
\ENSURE{A set of points on 3D front-view lanes ground truth $\mathcal{A}_{f}$.}

\STATE Compute transformation matrix $\bm{M}$ and its inverse $\bm{M}^{-1}$: $\bm{M}  \leftarrow \bm{K} \bm{T}$, $\bm{M}^{-1} \leftarrow Inverse(\bm{M})$.
\STATE Convert $\mathcal{A}_{sv}$ to generate lanes $\mathcal{A}_i$ in the front-view image space: $\mathcal{A}_{i} \leftarrow \bm{M} \mathcal{A}_{sv}$.
\STATE Filter $\mathcal{A}_{i}$ based on image dimensions: $\mathcal{A}_{i} \leftarrow RangeFilter(\mathcal{A}_{i}, {h}, {w})$.
\STATE Transform $\mathcal{A}_{i}$ to 3D front-view space: $\mathcal{A}_{f} \leftarrow {\bm{M}^{-1} \mathcal{A}_{i}  }$.
\STATE Return $\mathcal{A}_{f}$.
\end{algorithmic}
\end{algorithm}

\vspace{-0.1cm}
\subsection{Experiment Settings}
\textbf{Perception Range.} The proposed network can perform 3D lane detection tasks in both front-view and surround-view perspectives. To ensure fairness in benchmark evaluations, the visual ranges in the datasets align with current state-of-the-art methods. Specifically, in the Openlane dataset, the 3D-space range for 3D front-view lane detection tasks is set as $[-30m, 30m] \times [3m, 103m] \times [-10m, 10m]$ along the X, Y, and Z axes. Meanwhile, for the Argoverse2 dataset, the ranges are $[-15.0m, 15.0m]$ (X-axis), $[-30.0m, 30.0m]$ (Y-axis), and $[-2.0m, 2.0m]$ (Z-axis) for 3D surround-view lane detection. The Argoverse2$^{\dag}$ dataset's perception ranges for 3D front-view lane detection are defined similarly but with $[0.0m, 30.0m]$ for the Y-axis. Moreover, we set the divider number of the x, y, and z axes to 100, 50, and 4, respectively, to segment the 3D grid into multiple voxels.

\textbf{Point Number.} The Openlane dataset primarily offers straight lanes extending forward. We represent a lane curve using 20 key points and 5 control points. For the Argoverse2 and Argoverse2$^{\dag}$ datasets, which include complex lane shapes like closed-loop or loop, we use 20 key points and 10 control points for representation.

\textbf{Hyper Parameter.} For the Openlane dataset, our network scales input images by $0.8$ times and performs category classification on $17$ classes (different from Persformer, we do not merge left and right road edges into a single class). For Argoverse2 and Argoverse2$^{\dag}$ datasets, it scales images by $0.3$ times and classifies the lane line into four classes (dividers, pedestrian crossings, boundaries, and background classes). The training leverages the AdamW \cite{loshchilov2017decoupled} optimizer, with a learning rate of $6e^{-4}$ and weight decay of $0.01$. Optimal results were achieved in $36$ epochs using a single A800 GPU with a batch size of $8$. 

\textbf{Metric.} We adopt the same evaluation protocol used in the original Openlane's work that proposes to separate the detection accuracy from the geometric estimation accuracy. The official metric, F-Score, is derived as:
\vspace*{-0.3 \baselineskip}
\begin{equation}
F_{score} = \frac{2 \times Precision \times Recall}{Precision + Recall},
\vspace*{-0.3 \baselineskip}
\end{equation}
\vspace*{-0.3 \baselineskip}
\begin{equation}
Precision = \frac{TP}{TP + FP},
Recall = \frac{TP}{TP + FN}.
\vspace*{-0.3 \baselineskip}
\end{equation}

Given the predicted $L$ number of lane lines from $\bm K$, we compute the confidence of each lane line by category. Matches with the ground truth are considered when confidence exceeds 0.25. For details on Category Accuracy, readers are directed to the Openlane evaluation protocol.

Additionally, average precision (AP) evaluates the 2D map quality. Using the Chamfer distance $D_{Chamfer}$, matches between predictions and ground truth are determined. We compute the $AP_{\mu}$ under several $D_{Chamfer}$ thresholds $(\mu \in T, T = {0.5, 1.0, 1.5})$, averaging them to as the final AP metric:
\vspace*{-0.3 \baselineskip}
\begin{equation}
AP = \frac{1}{T} \sum_{\mu \in T} AP_{\mu}.
\vspace*{-0.3 \baselineskip}
\end{equation}

It is important to note that to ensure fairness in comparison, all the above metrics are calculated using only the interpolation curve as a benchmark, while the B\'{e}zier curve is used solely for model training.

\section{Results and Ablation}
To evaluate the efficacy of our methods, we examined the modeling accuracy and 2/3D performance across three datasets. We conducted comparative analyses with several 3D front-view techniques, namely 3D-LaneNet, Gen-LaneNet, Persformer, and CurveFormer. Notably, to the best of our knowledge, our network is the pioneering model for 3D surround-view lane prediction. To validate its performance in this domain, we mapped the 3D results onto a 2D plane and compared them with leading 2D surround-view lane detection techniques.

This section begins by evaluating various lane modeling methods. It then proceeds to assess the advantages of our approach in front-view 3D lane detection, employing both simple (Openlane) and complex (Argoverse2$^{\dag}$) datasets. Finally, we will evaluate the performance of 3D surround-view lane detection using an Argoverse2 dataset. It is important to note that anchor-based methods perform well on the Openlane dataset, but show poorer performance when dealing with complex curves present in the front-view samples from Argoverse2$^{\dag}$.

\begin{table*}[!b]
\vspace{-0.4cm}
\footnotesize
\centering
\caption{Performance comparison with other state-of-the-art 3D lane methods on Openlane benchmark.}
\label{Tab01}
\begin{threeparttable} 
\begin{tabular}{lccccccc}

\toprule
Method		&All		&Up\&Down		&Curve		& Extreme Weather		&Night		&Intersection		&Merge\&Split		\\
\midrule
3D-LaneNet \cite{garnett20193d}	&44.1	&40.8			&46.5		&47.5				&41.5		&32.1			&41.7			\\
Gen-LaneNet \cite{guo2020gen}	&32.3	&25.4			&33.5		&28.1				&18.7           	&21.4			&31.0			\\
PersFormer \cite{chen2022persformer}	&50.5	&42.4          		&55.6          	&48.6          			&46.6		&40.0			&50.7			\\
CurveFormer \cite{bai2023curveformer}	&50.5	&45.2			&56.6		&47.9				&49.1		&42.9			&45.4			\\
\hline\hline
Ours  		&55.7\cellcolor{lightgray}	&48.7\cellcolor{lightgray}			&60.3\cellcolor{lightgray}		&50.1\cellcolor{lightgray}				&52.1\cellcolor{lightgray}		&51.3\cellcolor{lightgray}			&51.9\cellcolor{lightgray}         		 \\
\bottomrule
\end{tabular}
\begin{tablenotes} 
	\item Evaluation on the F-Score metric. F-Score: higher is better.
\end{tablenotes} 
\end{threeparttable} 
\vspace{-0.4cm}
\end{table*}

\begin{figure}[t]
\centering{
\subfigure[Simple Curves. This frame represents a simple curve where no lines exhibit a cosine value exceeding 45 degrees in the X-Y plane. On the left sample, all three modeling approaches accurately replicate the annotated data. Conversely, on the right, the B\'{e}zier Curve exhibits notable oscillations when modeling intersecting curves.]{
\label{Fig4.sub.1}
\includegraphics[width=3.5in]{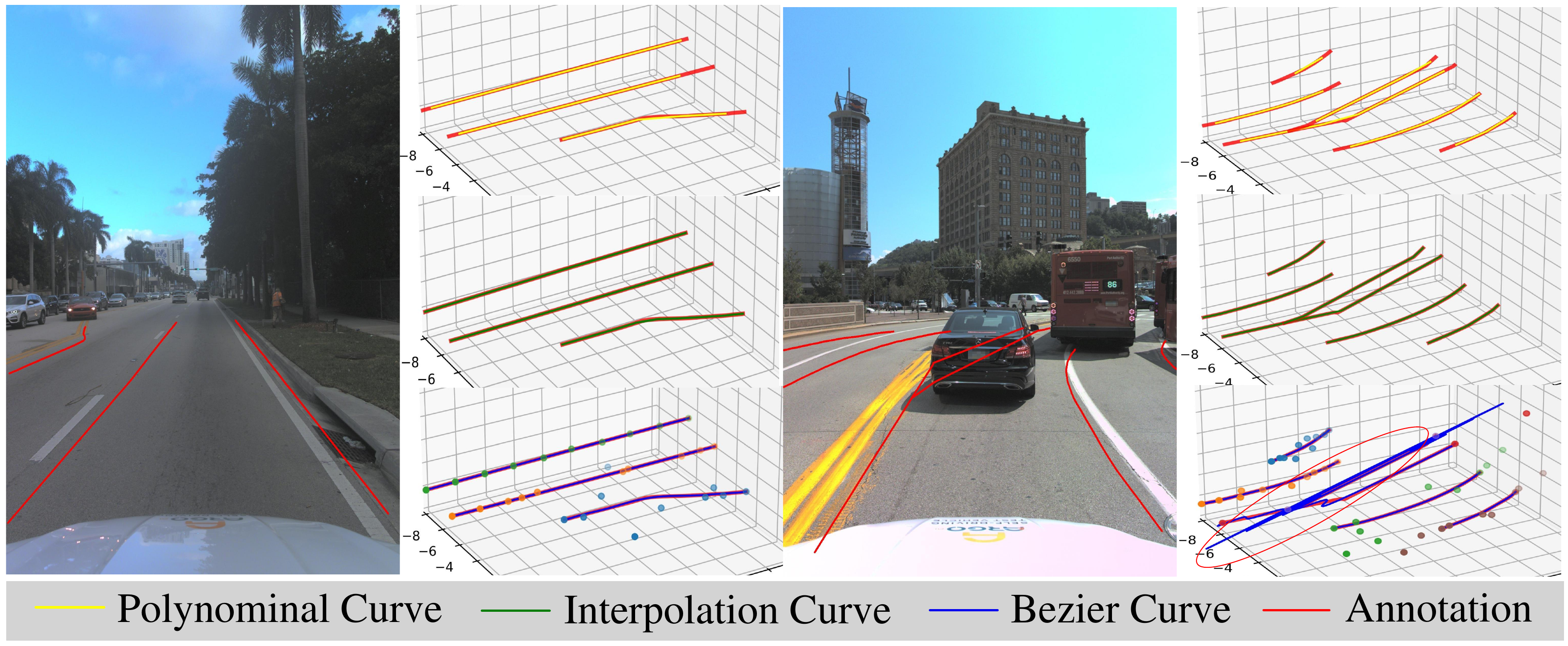}}
\subfigure[Complex Curves. Represented in this frame is a complex curve containing a line with a cosine value surpassing 45 degrees at various points on the X-Y plane.
We contend that the Polynomial Curve is unable to fit complex U-shaped curves in this condition, whereas the B\'{e}zier Curve and Interpolation Curve fit the annotated data greatly. Moreover, in the left sample, the B\'{e}zier Curve is smoother and more accurate, and in the right sample, the higher-order B\'{e}zier Curve shows slight oscillations.]{
\label{Fig4.sub.2}
\includegraphics[width=3.5in]{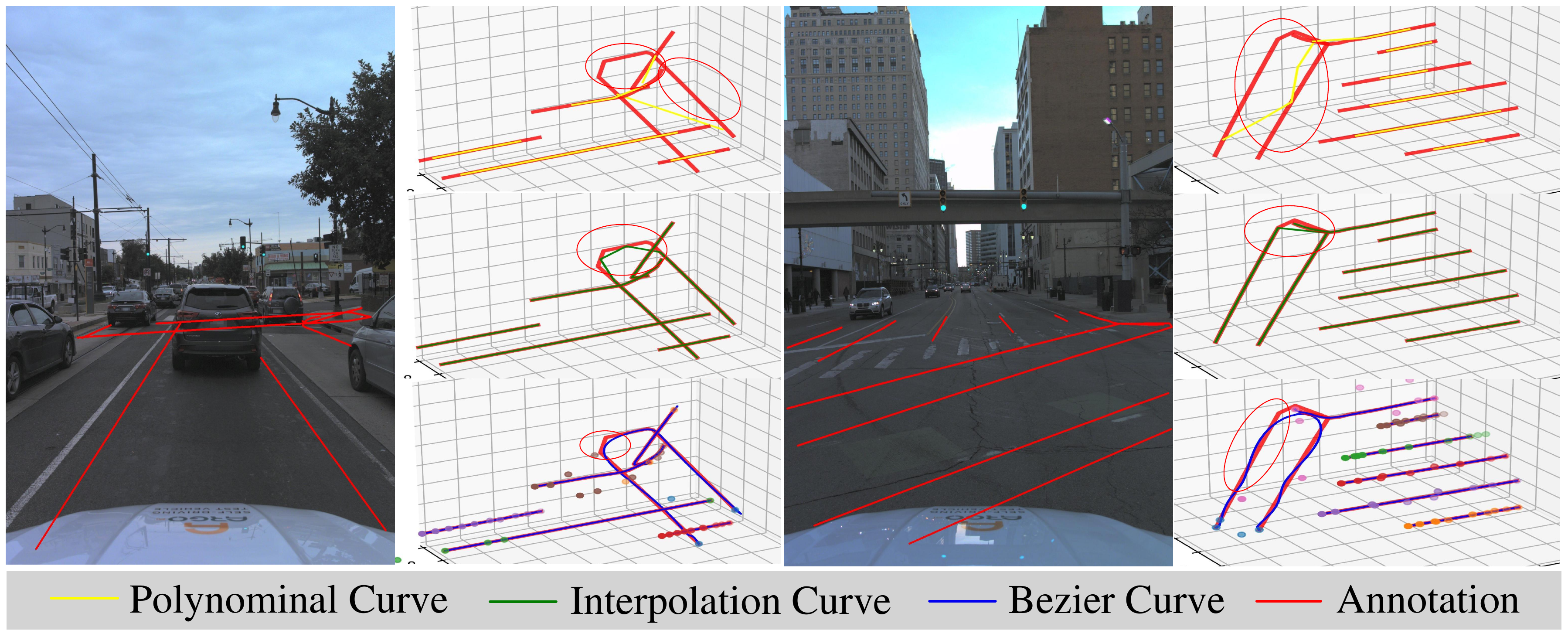}}
\caption{A comparison of the lanes modeling capabilities of the three modeling methods reveals the Polynomial Curve's suboptimal performance. In contrast, both the Interpolation Curve and B\'{e}zier Curve spotlight distinct advantages.}
\label{fig_44}}
\vspace{-0.4cm}
\end{figure}

\vspace{-0.1cm}
\subsection{Precision of Modeling}
Three modeling methods, namely the Polynomial Curve, Interpolation Curve, and B\'{e}zier Curve, are visualized as shown in Fig.~\ref{fig_44}. Datasets are classified into simple and complex categories depending on the presence of curves in the X-Y plane with cosine values surpassing 45 degrees. Notably, this categorization will be employed for the Argoverse2$^{\dag}$ dataset in subsequent tests. We observed that when handling complex curves, as illustrated in Fig.~\ref{Fig4.sub.2}, the relatively low-order Polynomial Curve was unsuccessful in accurately modeling the annotated data. This highlights the necessity of creating Argoverse2$^{\dag}$, which compensates for the simplicity of the Openlane dataset samples. The Interpolation Curve and B\'{e}zier Curve, to some extent, recreate real-world scenarios. We found that the Interpolation Curve handles straight lines with small curvatures effectively. However, when dealing with U-shaped curves, Fig.~\ref{Fig4.sub.2} reveals that the Interpolation Curve required more key points to achieve precise modeling. The B\'{e}zier Curve outperforms the Interpolation Curve in modeling straight and U-shaped curves. Still, it demonstrates strong oscillations when modeling Y-shaped curves, and higher-order B\'{e}zier also exhibits minor oscillations in the straight segment of the U-shaped curve.

Consequently, we propose the concept of Joint Lane Modeling, controlling the modeling effects of the B\'{e}zier Curve and Interpolation Curve through $\lambda, \alpha, \beta, \gamma$, and $\delta$. 

\vspace{-0.1cm}
\subsection{Front-View Detection Results on Openlane}
\textbf{Qualitative Comparisons.} As shown in Table~\ref{Tab01}, we compare our results with CNN-based 3D lane detection and Transformer-based 3D lane detection. Experimental results verify that our method outperforms the previous benchmarked approaches on the validation dataset, and achieves the top F-Score across different scenarios. By directly calculating the 3D coordinates of lane points, in contrast to anchor-based methods which adjust the Z and X axes while fixing the Y axis, we notice a significant improvement in the Curve, Intersection, and Merge \& Split categories. These categories involve numerous complex lines, and our method shows the inherent superiority over anchor-based approaches.

\begin{figure*}[t]
\centering
\includegraphics[width=6.5in]{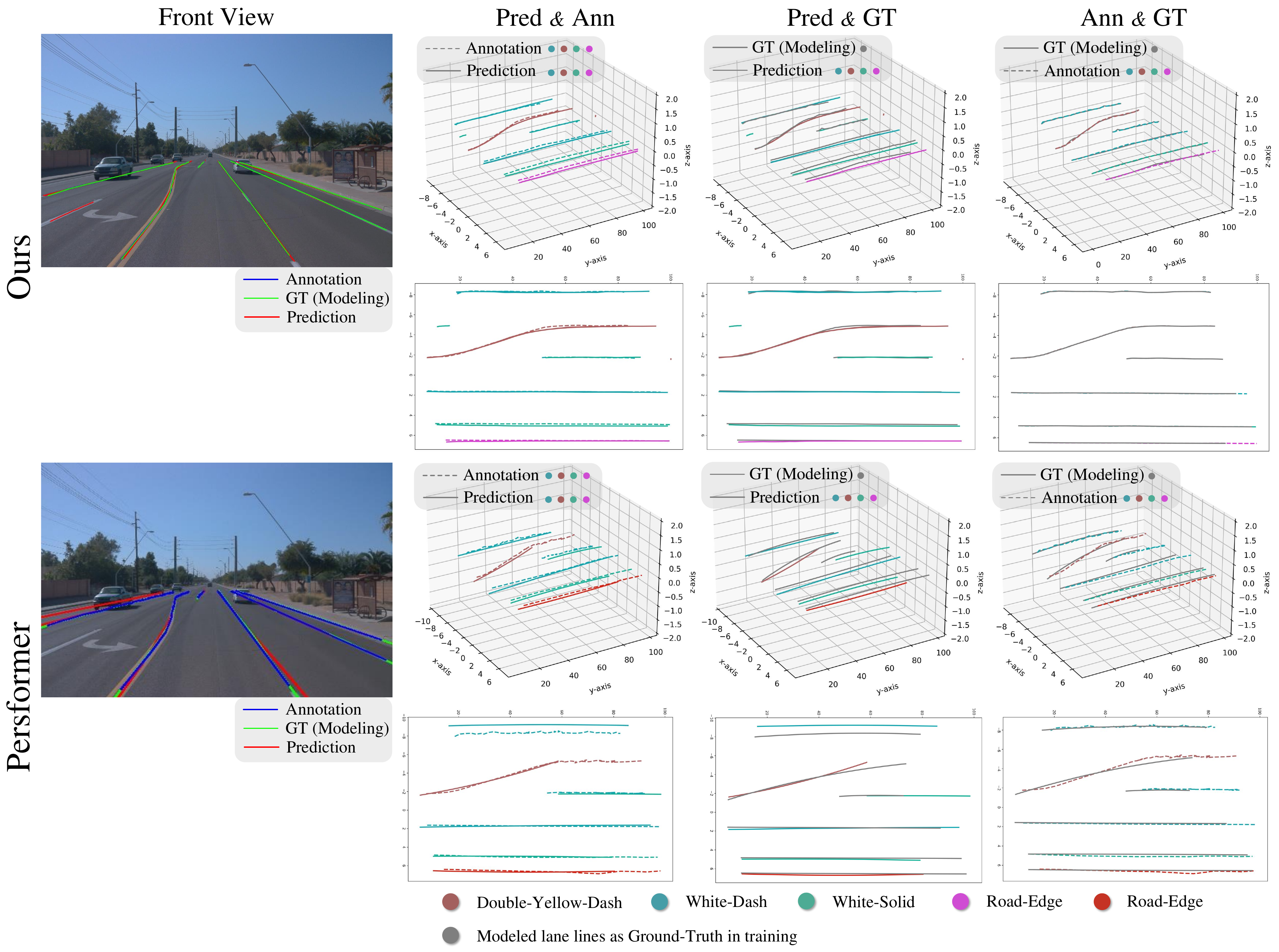}
\caption{Visualization comparison of 3D front-view lane results. It illustrates predictions versus annotated data (Pred\&Ann), predictions versus modeling ground truth (Pred\&GT), and annotated data versus modeling ground truth (Ann\&GT) from both 3D and 2D perspectives. The Pred\& Ann comparison highlights the network's predictive capability, while the Pred\&GT comparison emphasizes precision. The Ann\&GT comparison evaluates how well the network's modeling mirrors real-world conditions. Lastly, projecting the prediction results into image space further validates their effectiveness. Our network surpasses the Performer in terms of line position accuracy and the rendition of complex lines.}
\label{fig_5}
\vspace{-0.4cm}
\end{figure*}

\textbf{Quantitative Comparisons.}
As shown in Fig.~\ref{fig_5}, a visual comparison is conducted between our outcomes and the traditional anchor-based method. We consider annotated data to be the most accurate and authentic representation of the real-world lane line positions, while modeling ground truth may introduce some errors (compare it to the annotated data) to influence the accuracy of the prediction. Therefore, we performed comparisons of annotated data (Ann), modeling ground truth (GT), and prediction (Pred) results in both 2D and 3D (2D results are obtained by projecting 3D data onto the X-Y plane). 

In the 2D and 3D comparison plots of \textbf{Ann\&GT}, we observed that Persformer using Polynomial Curve can well fit the linear extension trend (forward), but it fails to describe the details of local curvature in curves precisely and does not perform well in fitting the brown curve. In contrast, our method combines the Interpolation Curve and B\'{e}zier Curve to describe a curve (we only display the fitting of the Interpolation Curve in the plot, and use the B\'{e}zier Curve's control points during result prediction) that fits the annotated data accurately and restores the distribution of real-world lanes, even achieving accurate results when fitting the brown curve. In \textbf{Pred\&GT}, we found that the prediction results strive to reproduce the modeling ground truth, moving closer to the perceived real-world situation due to deviations between the modeling lanes in Persformer and the actual situation.

Despite the loss constraints that make the predictions closer to the modeling ground truth, if there is a significant deviation between the modeling ground truth and annotated data, the prediction results cannot accurately reflect the real-world lane line positions, as shown in \textbf{Pred\&Ann}. The advantages of our method are evident in predicting complex curve types of lanes, such as the brown lanes.

Moreover, when projecting Ann, Pred, and GT into the image space, it is evident that our methodology ensures remarkably precise lane detection.

\begin{table}[t]
\footnotesize
\centering
\caption{Performance comparison with other state-of-the-art 3D lane methods on Argoverse2$^{\dag}$.}
\label{Tab03}
\begin{threeparttable} 
\begin{tabular}{lccc}

\toprule
Method		&Simple Curve	&Complex Curve\\
\midrule
Persformer \cite{chen2022persformer}	&46.0		&19.8\\
Ours			&47.7\cellcolor{lightgray}		&32.5\cellcolor{lightgray}\\
\bottomrule
\end{tabular}
\end{threeparttable} 
\vspace{-0.4cm}
\end{table}

\vspace{-0.1cm}
\subsection{Front-View Detection Results on Argoverse2$^{\dag}$}
To verify the front-view 3D lane detection capability of our method on complex-shaped lanes, we created the Argoverse2$^{\dag}$ based on Argoverse2. Compared to Openlane, Argoverse2$^{\dag}$ provides more complex-shaped lanes, including loops and closed loops, to better represent real-world scenarios. To better assess our method and Persformer's performance on Argoverse2$^{\dag}$, we divided this dataset into two categories: Simple Curve scenes and Complex Curve scenes. We defined any curve in a scene with a cosine value greater than $45$ degrees on the X-Y planes as a Complex Curve scene, while Simple Curve scenes mainly consist of straight lines extending forward.

\begin{table*}[!ht]
\footnotesize
\centering
\caption{Performance comparison with other state-of-the-art surround-view lane detection methods on Argoverse2.}
\label{Tab02}
\begin{threeparttable} 
\begin{tabular}{llcccccc}
\toprule
\multirow{2}{*}{Method}		& \multicolumn{4}{|c|}{2D Benchmark}                                                            & \multicolumn{2}{c}{3D Benchmark}              \\
	& \multicolumn{1}{|l}{$\rm AP_{ped}$} 		&$\rm AP_{divider}$		&$\rm AP_{boundary}$		&\multicolumn{1}{l|}{$\rm mAP$} 	&Category Accuracy		&F-Score                       \\
\midrule
HDMapNet \cite{li2022hdmapnet}  		                 & \multicolumn{1}{|c}{13.1} 		& \multicolumn{1}{c}{5.7} 		&\multicolumn{1}{c}{37.6} 		&\multicolumn{1}{c|}{18.8} &-\					&-\ 					\\
VectorMapNet	\cite{liu2023vectormapnet}	                 & \multicolumn{1}{|c}{38.3} 		& \multicolumn{1}{c}{36.1} 		&\multicolumn{1}{c}{39.2} 		&\multicolumn{1}{c|}{37.9} &-\					&-\ 					\\
MapTR-tiny \cite{liao2022maptr}  		                 & \multicolumn{1}{|c}{57.0\cellcolor{lightgray}} 		& \multicolumn{1}{c}{45.7} 		&\multicolumn{1}{c}{54.8\cellcolor{lightgray}} 		&\multicolumn{1}{c|}{52.5\cellcolor{lightgray}} &-\					&-\ 					\\

\hline\hline
Ours   			                 & \multicolumn{1}{|c}{33.6} 		& \multicolumn{1}{c}{49.9\cellcolor{lightgray}} 		&\multicolumn{1}{c}{50.8} 		&\multicolumn{1}{c|}{44.7} 	&\multicolumn{1}{c}{80.9\cellcolor{lightgray}}		&\multicolumn{1}{c}{44.5\cellcolor{lightgray}} 					\\

\bottomrule
\end{tabular}
\end{threeparttable} 
\vspace{-0.2cm}
\end{table*}

Persformer, as a representative work for front-view 3D lane detection, although the Polynomial Curve used in Persformer accurately fits straight lines, it fails to model complex curves. Our network significantly outperforms Persformer in complex curve scenarios, as shown in Table~\ref{Tab03}, our method has achieved remarkable performance in both scenarios. Even in the case of predicting simple curve scenes that are mostly straight lines, our method outperforms Persformer by 1.7 on  Argoverse2$^{\dag}$.

\begin{figure}[t]
\centering
\includegraphics[width=3.5in]{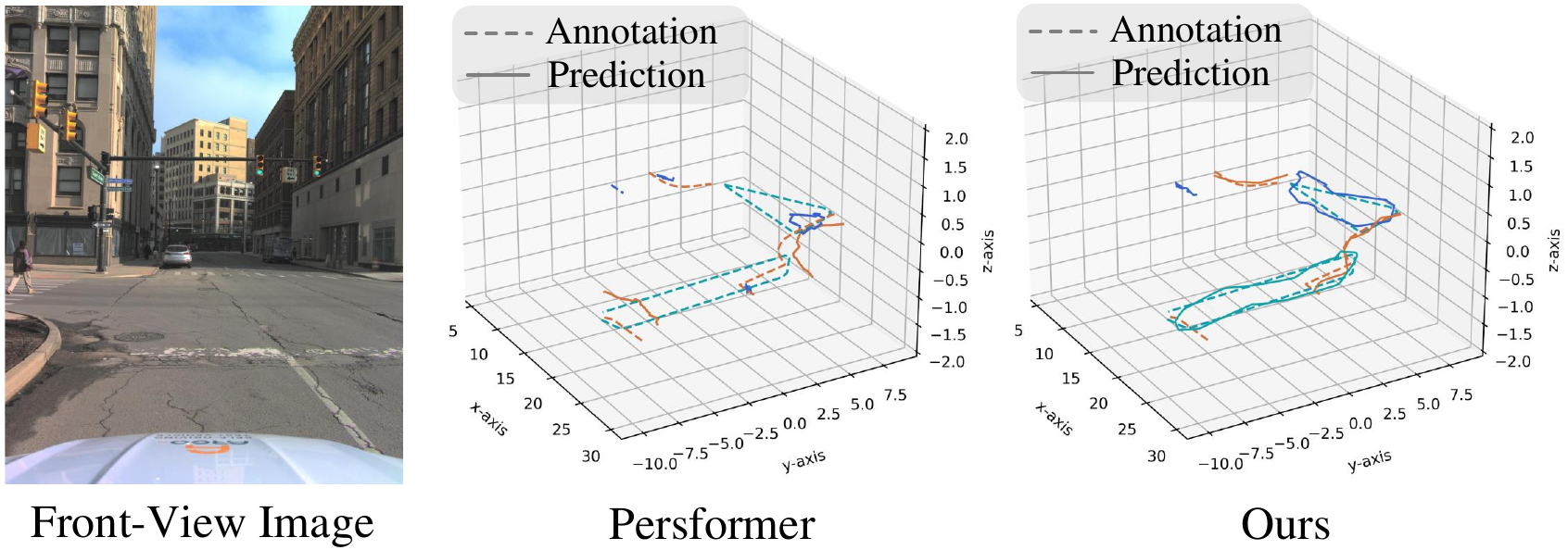}
\caption{Illustration of the significant advantage of our network in scenarios with complex curves.}
\label{fig_10}
\vspace{-0.4cm}
\end{figure}

In Fig.~\ref{fig_10}, our network not only accurately predicts the category of lanes but also captures their true positions in the real world. In contrast, Persformer has difficulty predicting closed-loop lane lines, but our joint modeling approach offers a closer reflection of real-world scenarios.

\begin{figure*}[t!]
\centering
\includegraphics[width=6.6in]{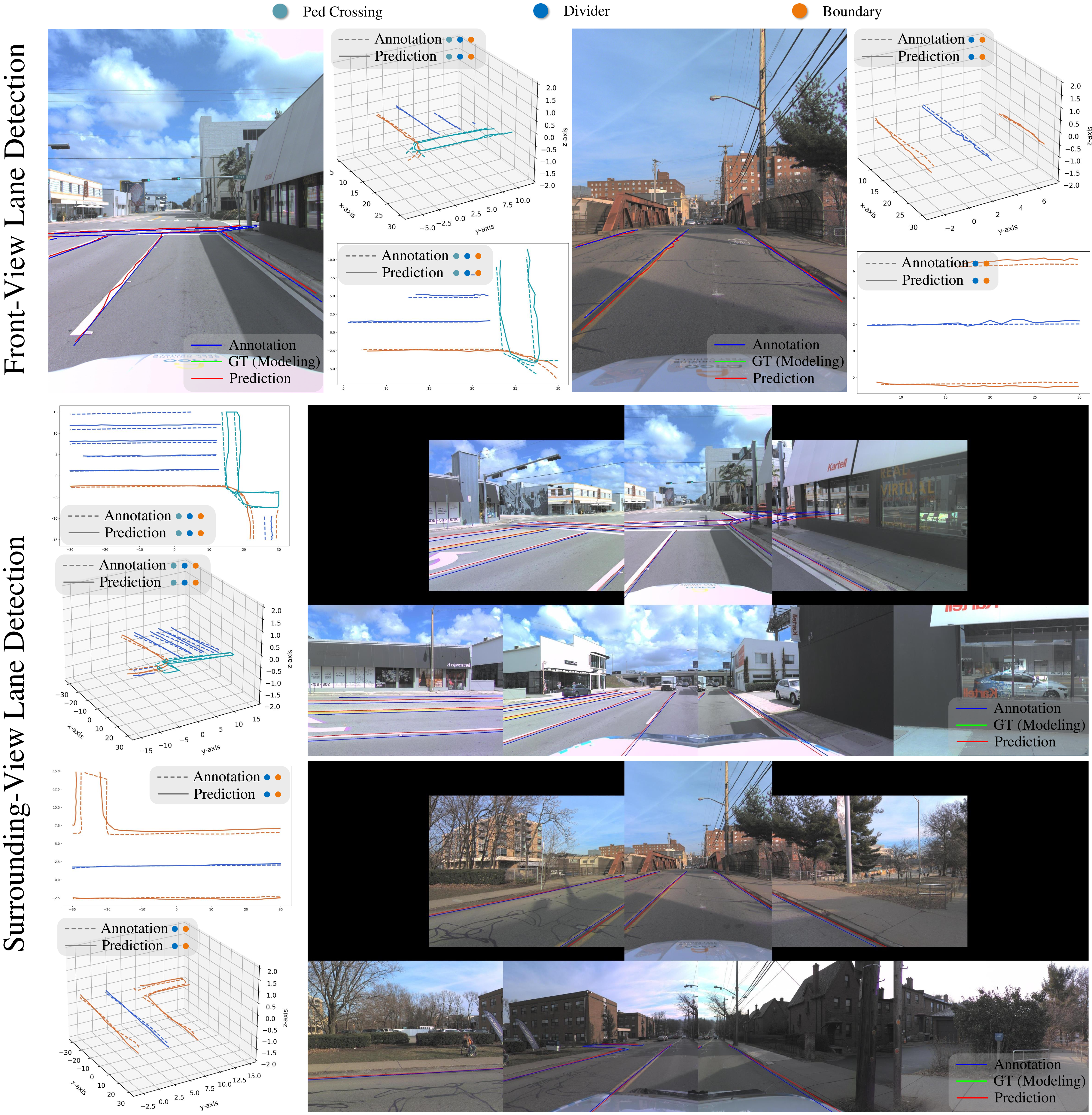}
\caption{Prediction results of the front-view 3D lanes in the Argoverse2$^{\dag}$ dataset and the surround-view 3D lanes in the Argoverse2 dataset. We compare the predicted results with the annotated data, rather than the modeling results. The colors indicate the categories of the lane lines.}
\label{fig_6}
\vspace{-0.4cm}
\end{figure*}

\vspace{-0.1cm}
\subsection{Surround-View Detection Results on Argoverse2}
As the first method achieving 3D surround-view lane detection, our approach draws inspiration from state-of-the-art 2D surround-view lane detection methods, elevating the prediction results to 3D space. However, we have observed that existing 2D surround-view lane detection methods, especially those pioneering in transforming the surround-view to a bird's-eye-view along with their unique detection techniques, cannot be easily modified by simply substituting their modules with our approach for 3D detection comparison. Therefore, for fair evaluation, we project the predicted results onto the X-Y plane and compare them with HDMapNet \cite{li2022hdmapnet}, VectorMapNet \cite{liu2023vectormapnet}, and MapTR-tiny \cite{liao2022maptr}. We perform comparisons based on the AP metric for the ped crossing, divider, and boundary classes. The three comparison methods were trained on Argoverse2 with the official provided iteration numbers until fitting, and the results are presented in Table~\ref{Tab02}.

Comparison in Table~\ref{Tab02}, we observe that our method surpasses the current best 2D surround-view lane detection method in the divider class, and significantly outperforms VectorMapNet in both the divider and boundary classes. Taking into consideration the mAP metric comprehensively, our method demonstrates competitive performance in 2D surround-view lane detection.

We visualize the results of 3D surround-view lane detection in Fig.~\ref{fig_6}. From the predicted results in 2D, 3D, and image spaces, our method accurately detects the surrounding lanes and classifies them correctly.
\vspace{-0.1cm}

\subsection{Ablation Study}
\textbf{B\'{e}zier Control Point Number.} The number of B\'{e}zier control points and key points jointly affect the experimental results. MapTR verifies the impact of the number of key points on the network's performance. Its surround-view BEV lane detection network achieves optimal performance when set the key points number to $20$. Therefore, in order to validate the influence of the number of B\'{e}zier control points on the network's performance, experiments are conducted with a default setting of $20$ key points as the optimal parameter value. The number of B\'{e}zier control points is set to $0$, $5$, $10$, and $15$. The impact of the control point number is evaluated using the Category Accuracy and F-Score metrics. The experiments are conducted on the Argoverse2 dataset, and the results are illustrated in Table~\ref{Tab04}.

\begin{table}[t]
\footnotesize
\centering
\caption{Comparison of performance with different numbers of B\'{e}zier control points on Argoverse2.}
\label{Tab04}
\begin{threeparttable} 
\begin{tabular}{ccc}
\toprule
C-Pt. Num.	&Catagory Accuracy	&F-Score	\\
\midrule
0			&78.0			&42.3\\
5			&80.8			&43.9\\
10			&80.9\cellcolor{lightgray}          &44.5\cellcolor{lightgray} \\
15			&80.9          		&44.3\\
\bottomrule
\end{tabular}
\end{threeparttable} 
\vspace{-0.2cm}
\end{table}

\begin{figure}[!ht]
\centering
\includegraphics[width=3.2in]{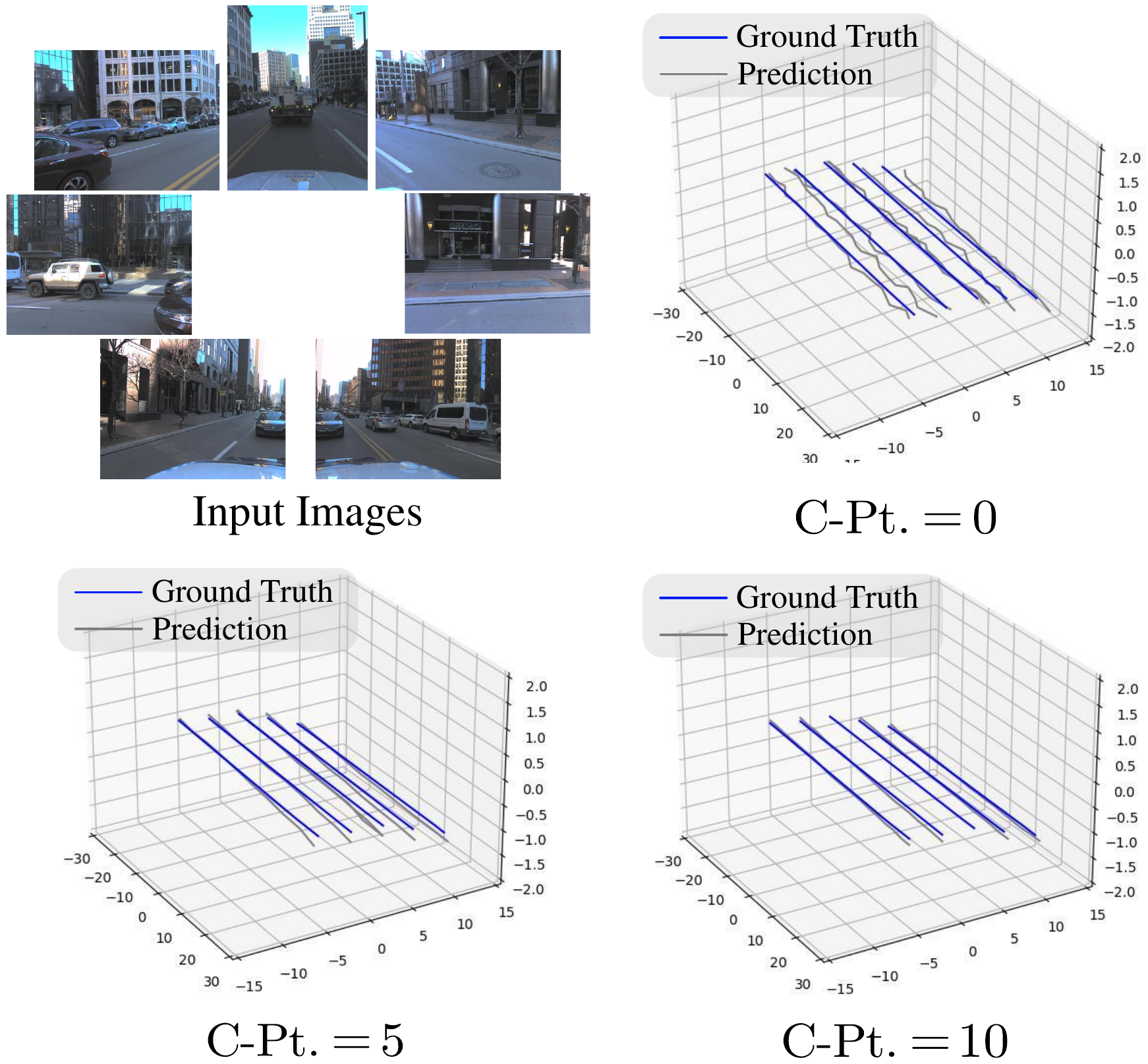}
\caption{Impact of the number of control points on predicting lanes.}
\label{fig_8}
\vspace{-0.4cm}
\end{figure}

By comparing the data, we observed that both the results obtained with the control point numbers set to $5$, $10$, and $15$ outperform the results without any control points being applied. This suggests that employing B\'{e}zier control points is advantageous in enhancing the precision of detection outcomes. Fig.~\ref{fig_8} shows that the number of control points affects the smoothness of the predicted curve. Moreover, the experiment achieves the optimal value when the number of B\'{e}zier control points is set to $10$ on Argoverse2.

\textbf{Key Point Number.} The number of key points determines the ability to represent the curvature of the lanes. When highly curved lanes are represented with only a few key points, this can lead to discontinuities and a loss of overall smoothness. After determining the optimal number of B\'{e}zier control points, we proceed to examine the influence of the number of key points by setting them to $10$, $20$, and $40$. The experiments were conducted using the Argoverse2 dataset, and the results are illustrated in Table~\ref{Tab05}.

\begin{table}[t]
\footnotesize
\centering
\caption{Comparison of performance with different numbers of key points.}
\label{Tab05}
\begin{threeparttable} 
\begin{tabular}{ccc}
\toprule
K-Pt. Num.	&Catagory Accuracy	&F-Score	\\
\midrule
10			&79.7			&43.0\\
20			&80.9\cellcolor{lightgray}			&44.5\cellcolor{lightgray}\\
40			&80.5          		&43.3\\
\bottomrule
\end{tabular}
\end{threeparttable} 
\vspace{-0.2cm}
\end{table}

The table data indicates that too many key points can also impact the network's accuracy, with the best performance achieved when using $20$ key points.

\textbf{Robustness to Weather Conditions.}
The weather has a pronounced impact on sensor performance, and the image quality from onboard cameras plays a pivotal role in network inference. It is imperative to have a robust neural network to navigate the challenges presented by diverse weather conditions. As shown in Table~\ref{Tab01}, an analysis of the Openlane dataset reveals marked discrepancies in network performance during Extreme Weather and Night conditions. Significantly, except for 3D-LaneNet, all methods experience a dip in accuracy in such corner cases. Gen-LaneNet displays the most significant drop, while CurveFormer maintains the most consistent performance, with only a slight decrease in accuracy.

\begin{table}[]
\footnotesize
\centering
\caption{Comparison of performance in different weather on Argoverse2.}
\label{Tab07}
\begin{threeparttable} 
\begin{tabular}{cccccc}
\toprule
Weather		&Normal			&Sunny	&Rainy	&Foggy	&Night	\\
\midrule
F-Score		&44.5			&47.5	&42.3	&42.0	&39.1	\\

\bottomrule
\end{tabular}
\end{threeparttable} 
\vspace{-0.5cm}
\end{table}

In the Argoverse2 dataset, samples are not directly categorized by weather conditions. To address this, we handpicked samples representing four distinct weather scenarios: Sunny, Rainy, Foggy, and Night. Each weather-specific sample contains 100 image sets. We carried out comparative studies across these four categories. As shown in Table~\ref{Tab07}, the prediction results for the Normal sample are sourced from Table~\ref{Tab02}. Here, Normal refers to a sample representing various weather conditions. Analyzing Table~\ref{Tab07}, it is evident that the clarity of sunny days results in enhanced performance during Sunny conditions, surpassing the usual benchmarks (Normal). On the flip side, due to limited visibility at night, the prediction accuracy for the Night category is the lowest, showing a decline of approximately 12.1\% compared to the Normal benchmark.

\textbf{Best-Performing Decoder Layer Number.} The decoder contains $\ell$ layers of transformer decoder. This ablation experiment aims to identify the optimal number of layers. Table~\ref{Tab06} presents the results from each layer's output in the decoder. It is observed that the best performance in surround-view 3D lane detection is achieved when the number of layers is set to $6$.

\begin{table}[t]
\footnotesize
\centering
\caption{Comparison of performance with decoder layer number.}
\label{Tab06}
\begin{threeparttable} 
\begin{tabular}{ccc}
\toprule
Layer Num.	&Catagory Accuracy	&F-Score	\\
\midrule
1			&78.0			&36.6\\
2			&78.3			&38.1\\
3			&79.8          		&40.3\\
6			&80.9\cellcolor{lightgray}			&44.5\cellcolor{lightgray}\\
8			&80.8          		&42.0\\
\bottomrule
\end{tabular}
\end{threeparttable} 
\vspace{-0.2cm}
\end{table}

\begin{figure}[t]
\centering
\includegraphics[width=3.3in]{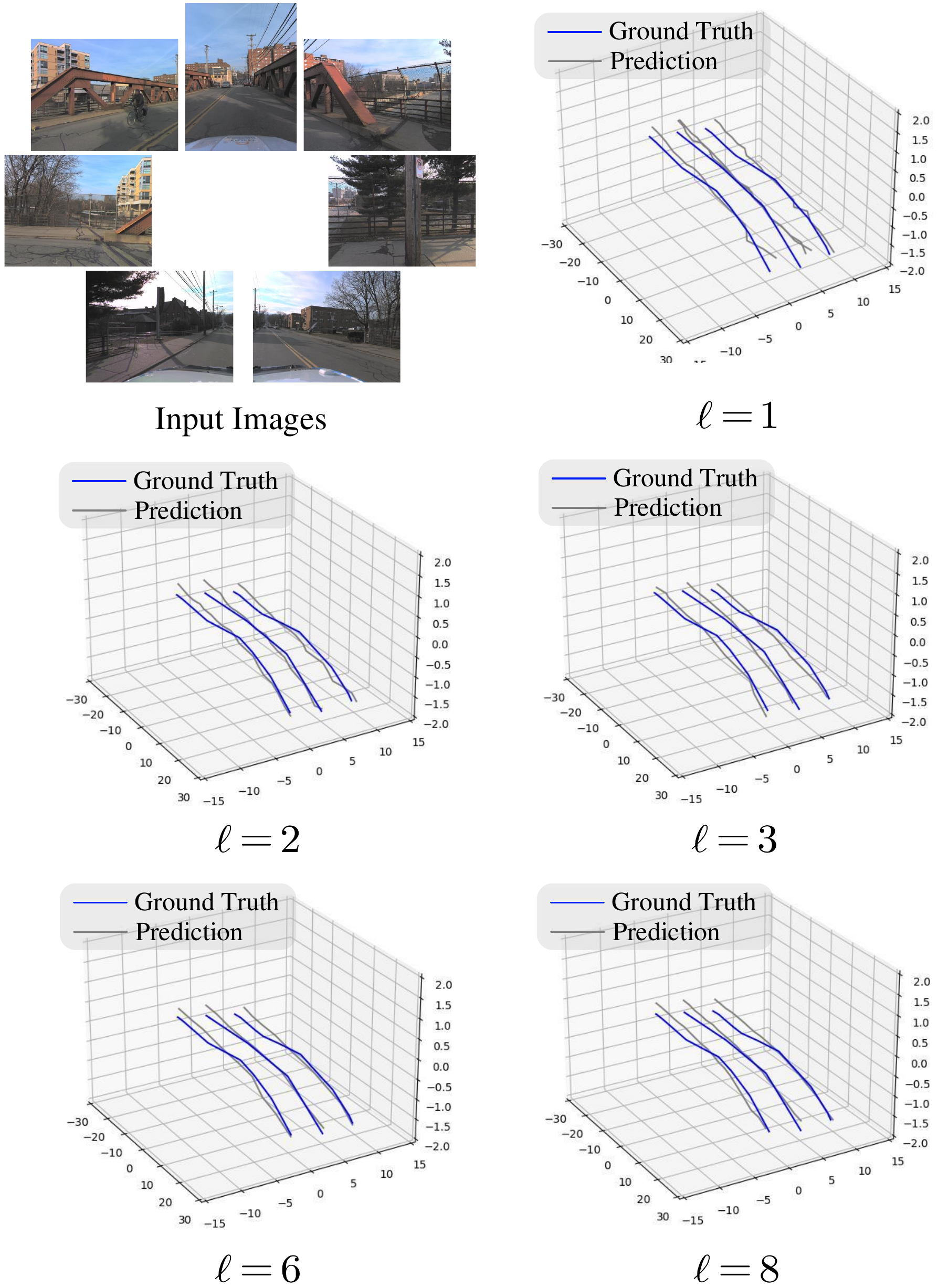}
\caption{Impact of the number of decoder layers on the prediction results.}
\label{fig_9}
\vspace{-0.4cm}
\end{figure}

Reviewing Fig.~\ref{fig_9}, we find that when $\ell=1$ the prediction performance is the worst, resulting in four predicted curves, with two corresponding to a single ground truth curve. The best performance is observed when $\ell=6$, while a declining trend in prediction performance is noted when $\ell=8$.

\section{Conclusions}
In this paper, we focus on precise and efficient lane modeling and the implementation of 3D lane construction from surround view images. We introduce a novel approach for front or surround-view 3D lane detection. To the best of our knowledge, this is the first paper that explores 3D lane line detection. Furthermore, our research offers a fresh perspective on the lane modeling challenge. We propose the concept of Joint Lane Modeling that can effectively handle complex curve modeling. Based on this joint modeling strategy, we introduce a GL-BK lane matching mechanism. This method stands as a comprehensive solution that leverages multiple features and mathematical models to ensure a precise match between predicted lane lines and their ground truths. Furthermore, we propose the 3D Spatial Encoder which lifts the front or surround-view features into the 3D space. Experimental results demonstrate that our approach achieves state-of-the-art results in front-view 3D lane prediction and competitive results in surround-view 3D lane prediction. We hope this paper will contribute to the advancement of 3D surround-view lane detection.

In future tasks, we will investigate the impact of corner cases on perception models and emphasize the construction of common neural network modules to enhance the framework's robustness against various edge cases.

\section{Acknowledgement}
We are grateful to Junjun Jiang for helpful discussions.

\bibliographystyle{IEEEtran}
\bibliography{references}
\vspace{-15 mm} 

\begin{IEEEbiography}[{\includegraphics[width=1in,height=1.25in,clip]{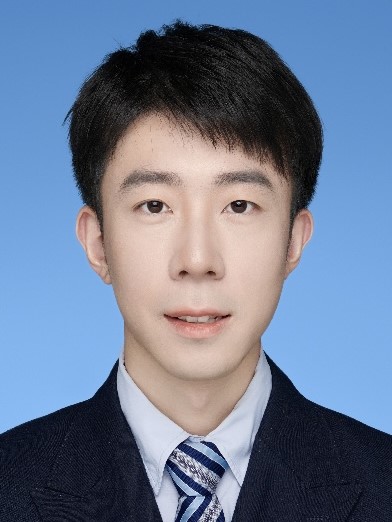}}]{Haibin Zhou}
received the B.S. degree from the Jiyang College of Zhejiang A\&F University, Zhejiang, China, in 2019. He is currently working toward a master's degree with the School of Hubei Key Laboratory of Intelligent Robot, Wuhan Institute of Technology, Wuhan, China. His research interests include computer vision, machine learning, and autonomous driving.
\end{IEEEbiography}
\vspace{-15 mm} 

\begin{IEEEbiography}[{\includegraphics[width=1in,height=1.25in,clip]{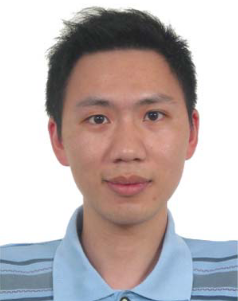}}]{Huabing Zhou}
(Member, IEEE) received the B.S. and M.S. degrees in computer science and technology from the Wuhan Institute of Technology, Wuhan, China, in 2005 and 2008, respectively, and the Ph.D. degree in control science and engineering from the Huazhong University of Science and Technology, Wuhan, China, in 2012. From 2009 to 2010, he was a Research Intern with the Chinese Academy of Surveying and Mapping. From 2018 to 2019, he was a Visiting Scholar with the Temple University, Philadelphia, PA, USA. He is currently a Professor at the School of Computer Science and Engineering, Wuhan Institute of Technology, Wuhan, China. His research interests include computer vision, remote sensing image analysis, intelligent robot, and autonomous driving.
\end{IEEEbiography}
\vspace{-15 mm} 

\begin{IEEEbiography}[{\includegraphics[width=1in,height=1.25in,clip]{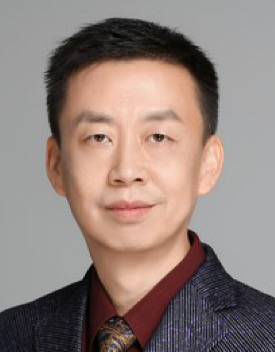}}]{Jun Chang}
received the B.S. degree and the Ph.D. degree from the School of Computer, Wuhan University, Wuhan, China. He was a Visiting Scholar at the Curtin University, Perth, Australia. He is involved in multiple national key research and development programs, major scientific and technological projects, and projects funded by the National Natural Science Foundation. He is currently a Lecturer at the Computer Science School, Wuhan University, Wuhan, China. His research interests include computer vision and machine learning.
\end{IEEEbiography}
\vspace{-15 mm} 

\begin{IEEEbiography}[{\includegraphics[width=1in,height=1.25in,clip]{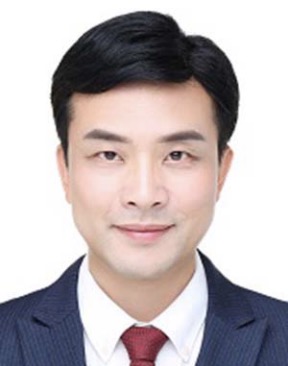}}]{Tao Lu}
(Member, IEEE) received the B.S. and M.S. degrees in computer applied technology from the School of Computer Science and Engineering, Wuhan Institute of Technology, Wuhan, China, in 2003 and 2008, respectively, and the Ph.D. degree in communication and information systems from the National Engineering Research Center For Multimedia Software, Wuhan University, Wuhan, China, in 2013. He is currently a Professor at the School of Computer Science and Engineering, Wuhan Institute of Technology, and a Research Member at Hubei Provincial Key Laboratory of Intelligent Robot. He was a Postdoc with the Department of Electrical and Computer Engineering, Texas A \& M University, College Station, TX, USA, from 2015 to 2017. His research interests include image/video processing, computer vision, and artificial intelligence.
\end{IEEEbiography}
\vspace{-15 mm} 

\begin{IEEEbiography}[{\includegraphics[width=1in,height=1.25in,clip]{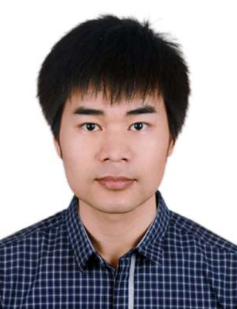}}]{Jiayi Ma}
(Senior Member, IEEE) received the B.S. degree in information and computing science and the Ph.D. degree in control science and engineering from the Huazhong University of Science and Technology, Wuhan, China, in 2008 and 2014, respectively. He is currently a Professor with the Electronic Information School, Wuhan University, Wuhan, China. He has authored or coauthored more than 200 refereed journal and conference papers, including IEEE TPAMI/TIP, IJCV, CVPR, ICCV, ECCV, \textit{etc}. His research interests include computer vision, machine learning, and robotics. He is identified with the 2019-2021 Highly Cited Researcher lists from the Web of Science Group. He is an Area Editor of \emph{Information Fusion}, and an Associate Editor of \emph{IEEE/CAA Journal of Automatica Sinica}, \emph{Neurocomputing}, \emph{Geo-spatial Information Science}, and \emph{Image and Vision Computing}.
\end{IEEEbiography}

\end{document}